\documentclass{article}
\usepackage{iclr2026_conference,times}

\usepackage{amsmath,amsfonts,bm}

\def\eqref#1{equation~\ref{#1}}

\def\1{\bm{1}}

\DeclareMathAlphabet{\mathsfit}{\encodingdefault}{\sfdefault}{m}{sl}
\SetMathAlphabet{\mathsfit}{bold}{\encodingdefault}{\sfdefault}{bx}{n}

\definecolor{errgrey}{gray}{0.45}
\definecolor{selbg}{RGB}{223,236,247}
\newcommand{\pmerr}[1]{\,{\color{errgrey}\scriptsize$\pm#1$}}

\usepackage[dvipsnames]{xcolor}  

\usepackage{hyperref}       
\hypersetup{
    colorlinks=true,
    linkcolor=MidnightBlue,
    citecolor=MidnightBlue,
    urlcolor=MidnightBlue,
    pdfborder={0 0 0},
    pdftitle={Quenched Ensemble Sampling},
    pdfauthor={David Yallup},
    pdfsubject={Samplers for Boltzmann distributions; nested sampling; flow matching},
    pdfkeywords={Sequential Monte Carlo, phase transitions, partition function, molecular sampling}
}

\usepackage{url}
\usepackage{xurl}
\usepackage{amsmath,amssymb}
\usepackage{booktabs}
\usepackage{graphicx}
\usepackage{xcolor}
\usepackage{caption}
\usepackage{xspace}
\usepackage{multirow}
\usepackage{colortbl}   
\usepackage{enumitem}
\usepackage{cleveref}
\crefname{equation}{Eq.}{Eqs.}
\Crefname{equation}{Eq.}{Eqs.}
\usepackage{algorithm}      
\usepackage{algpseudocode}  

\newcommand{\method}{QES\xspace}

\newcommand{\dlogz}{\mathrm{dlogz}}

\title{Quenched Ensemble Sampling}

\author{%
  David Yallup\thanks{dy297@cam.ac.uk} \\
  Kavli Institute for Cosmology Cambridge \\
  Institute of Astronomy, University of Cambridge, \\
  Madingley Road, Cambridge, CB3 0HA, UK
}

\iclrfinalcopy  

\begin{document}
\maketitle
\lhead{}

\begin{abstract}
Some of the sharpest challenges in sampling from the energy functions of physical systems arise at phase transitions, where the density of states changes abruptly and many sampling algorithms stall. Nested sampling is a particle method that traverses the density of states under a hard energy constraint and is known to be robust to such transitions, but its application in high dimension is limited by the difficulty of sampling under that constraint. In this work we introduce Quenched Ensemble Sampling, which generalises the hard constraint to a family of repulsive potentials at the energy boundary. This preserves the quenched path of monotonically decreasing energy while making the constrained target amenable to scalable gradient-based kernels. We demonstrate on synthetic models of phase transitions that our method estimates the marginal likelihood and draws posterior samples across a first-order transition where popular alternatives such as tempering fail. We apply the procedure to marginal likelihood estimation in Bayesian neural networks, enabling model comparison between network architectures. Finally, in a high-dimensional continuous lattice field theory, we show that this method traverses a first-order transition and estimates the partition function.

\end{abstract}

\section{Introduction}

Estimating the marginal likelihood is a central problem in Bayesian inference, and is particularly challenging for sampling algorithms. The marginal likelihood is defined as the integral, $Z = \int \mathcal{L}(x)\,\pi(x)\,\mathrm{d}x$, of a likelihood $\mathcal{L}$ over a prior $\pi$ on parameters $x$. In Bayesian inference this is the normalising constant of the posterior, and is central to Bayesian model comparison~\citep{llorente_marginal_2023}. Estimation of this integral is typically performed by introducing a series of interpolating distributions between the prior and posterior, where the integral can then be approximated by a series of ratios of normalising constants along this path. The most common approach is to use a temperature ladder, where the interpolating distributions are proportional to $\mathcal{L}^{\beta_k}(x)\,\pi(x)$
for a discrete schedule of inverse temperatures
$0 = \beta_0 < \beta_1 < \cdots < \beta_K = 1$. This ladder is traversed
sequentially in annealed importance sampling (AIS)~\citep{neal_annealed_2001}
and sequential Monte Carlo (SMC)~\citep{doucet_introduction_2001}, or held
concurrently as exchanging replicas in parallel
tempering~\citep{swendsen_replica_1986}. This makes the integral numerically tractable, and brings additional benefits for exploring the posterior landscape: at high temperatures the interpolating distributions flatten the energy barriers between modes, allowing transitions that are inaccessible at $\beta = 1$~\citep{syed_non-reversible_2022}.

Complementary to these approaches, it is possible to formulate a \emph{vertical} path through the density of states, rather than a \emph{horizontal} path through temperature~\citep{polson_vertical-likelihood_2015}. This is the basis of nested sampling~\citep{skilling_nested_2006}, which has been applied to a wide range of problems in physics~\citep{partay_efficient_2010,ashton_nested_2022}. This approach writes the marginal likelihood as an integral over the prior volume at monotonically increasing values of the likelihood; we review this construction in detail in~\Cref{sec:background}. The path of increasing likelihood is notably robust to first-order phase transitions, allowing the path of distributions to traverse regions of parameter space that hold a vanishing fraction of the prior mass, and which are therefore inaccessible to tempering methods. However, its practical implementation relies on sampling under a hard constraint, which degrades the efficiency of gradient-based MCMC kernels~\citep{kroupa_resonances_2025}. 

In this work we introduce \emph{Quenched Ensemble Sampling}, a method that preserves the attractive features of the path of increasing likelihood level sets, while incorporating a family of repulsive potentials at the hard constraint that allow the particles to efficiently move in high-dimensional spaces with standard gradient-based MCMC kernels~\citep{fearnhead_scalable_2025}. This path follows a trajectory of monotonically decreasing energy in an extended state space, which when implemented with an ensemble of particles, allows for a Sequential Monte Carlo sampler to be constructed that is robust to first order phase transitions. We formulate this procedure using motivations from statistical physics, highlighting the importance of this quenched path in the context of sampling from Boltzmann distributions, and demonstrate its performance on a range of problems, including synthetic models of phase transitions, a high-dimensional lattice field theory problem, and marginal likelihood estimation in Bayesian neural networks.

The key contributions of this work are:
\begin{itemize}
  \item We formulate an adaptive SMC sampler over a quenched family of distributions, and show that a family of repulsive potentials at the energy level defines a soft microcanonical ensemble, admitting gradient-based MCMC mutation kernels. 
  \item We demonstrate that this formulation connects classic nested sampling and tempering, with a family of potentials that interpolates between the two, generalising the nested sampling path to a broader family of distributions that hold a distribution of effective temperatures, rather than a hard constraint or isothermal ensemble (\Cref{app:nu}).
  \item We provide a practical implementation that incorporates a well-tuned local gradient-based Metropolis-adjusted Langevin Algorithm (MALA) MCMC kernel~\citep{roberts_exponential_1996} using the particle ensemble, and directly contrast with the tempering path on challenging black box high dimensional inference problems, which are inaccessible to standard nested sampling implementations.
\end{itemize}

\section{Background}
\label{sec:background}
To motivate the quenched path, we review the formulation of Bayesian inference
in the language of statistical mechanics, and the connection between the
marginal likelihood and the partition function. Connecting the statistical mechanics and
Bayesian inference perspectives, we write the canonical (Boltzmann)
density---in Bayesian terms, the tempered posterior---at inverse temperature
$\beta = 1/T$ as
\begin{equation}
  \rho_\beta(x) = \frac{1}{Z(\beta)}\, \mathcal{L}^\beta(x)\,\pi(x),
  \qquad
  Z(\beta) = \int \mathcal{L}^\beta(x)\,\pi(x)\,\mathrm{d}x\,,
\end{equation}
where the Boltzmann factor $e^{-\beta U(x)} = \mathcal{L}^\beta(x)$ is the
tempered likelihood, with energy $U(x) = -\log \mathcal{L}(x)$, and the prior
$\pi$ plays the role of the phase-space measure
(\Cref{tab:dictionary}). Samples from $\rho_\beta(x)$ constitute the
canonical ensemble, and the normalising constant $Z(\beta)$ is the partition
function, a central quantity in statistical mechanics; at $\beta = 1$ it is
the marginal likelihood, whose estimation underpins Bayesian model
comparison. In a statistical mechanics context, this integral is often
written in terms of the density of states,
\begin{equation}\label{eq:dos}
  g(E) = \int \delta\big(E - U(x)\big)\,\pi(x)\,\mathrm{d}x\,,
\end{equation}
the prior mass of configurations (microstates) per unit energy at $E$---the
analogue of the density of microstates, with the prior as the a priori
measure---giving
\begin{equation}
  Z(\beta) = \int \mathcal{L}^\beta(x)\,\pi(x)\,\mathrm{d}x
           = \int g(E)\, e^{-\beta E}\,\mathrm{d}E\,,
\end{equation}
which expresses the partition function as a Laplace transform of the density
of states. This is an integral on the energy axis, which motivates the vertical likelihood representation. The insight of nested sampling~\citep{skilling_nested_2006} lies
in defining the \emph{cumulative density of states},
$G(E) = \int_{-\infty}^{E} g(u)\,\mathrm{d}u$, the fraction of prior mass at
energies below $E$. Writing $E(G)$ for the generalized inverse of the
non-decreasing function $G(E)$, and substituting
$g(E)\,\mathrm{d}E = \mathrm{d}G$, gives
\begin{equation}\label{eq:evidence}
  Z(\beta) = \int g(E)\, e^{-\beta E}\,\mathrm{d}E
           = \int e^{-\beta E}\,\mathrm{d}G(E)
           = \int_0^1 e^{-\beta E(G)}\,\mathrm{d}G \,.
\end{equation}

This replaces the density of states $g(E)$, which is defined only through
infinitesimal energy shells and is therefore challenging to estimate
numerically, with its cumulative counterpart $G(E)$, a prior probability that
can be estimated directly by sampling. \Cref{eq:evidence} is the
vertical representation of the partition function: all dependence on the
$D$-dimensional parameter space is absorbed into the one-dimensional monotone
curve $E(G)$, and estimation reduces to resolving this curve at
exponentially small values of $G$.

\begin{table}[t]
  \centering
  \caption{Dictionary between Bayesian inference and statistical mechanics.
  The prior plays the role of the (usually implicit) a priori phase-space
  measure, and consequently enters the statistical-mechanics description only
  through the density of states, which is the prior mass per unit energy
  rather than a microstate count.}
  \label{tab:dictionary}
  \begin{tabular}{ll}
    \toprule
    \textbf{Bayesian inference} & \textbf{Statistical mechanics} \\
    \midrule
    Prior $\pi(x)\,\mathrm{d}x$ & Phase-space measure $\mathrm{d}\mu(x)$ \\
    Negative log-likelihood $-\log \mathcal{L}(x)$ & Energy $U(x)$ \\
    Tempered likelihood $\mathcal{L}^\beta(x)$ & Boltzmann factor $e^{-\beta U(x)}$ \\
    Tempered posterior $\rho_\beta(x)$ & Canonical (Boltzmann) distribution \\
    Marginal likelihood $Z = Z(1)$ & Partition function $Z(\beta)$ \\
    Likelihood level set $\{\mathcal{L}(x) > \lambda\}$ & Energy level set $\{U(x) < E\}$ \\
    Prior density of $-\log\mathcal{L}$ & Density of states $g(E)$ \\
    Cumulative prior mass $G(E)$ (prior volume $X$) & Cumulative density of states \\
    \bottomrule
  \end{tabular}
\end{table}

\subsection{Nested sampling and its limitations}
\label{sec:nested_sampling}

The nested sampling procedure was proposed by~\citet{skilling_nested_2006} as a generic particle method whose central computational requirement is efficient sampling from the constrained prior $\pi^*(x) = \pi(x)\,\mathbf{1}\{U(x) < E\}$. In practice, the most common approaches are to use rejection sampling with adaptive proposals~\citep{feroz_multinest_2009} in low dimensions and slice sampling~\citep{neal_slice_2003} as an MCMC mutation kernel that can scale to higher dimensions~\citep{handley_polychord_2015,yallup_nested_2026}. Problem-specific variants have made the method scalable to high dimensions: proximal operators in convex problems~\citep{mcewen_proximal_2023}, cluster moves in Potts models~\citep{murray_nested_2005}, and factorized blocked Gibbs samplers for hierarchical models~\citep{yallup_nested_2026-1}. However, as a general-purpose method, it has seen limited application beyond $D \sim 10^2$, where slice sampling becomes computationally expensive.

This limitation arises from the hard constraint: away from its boundary, the constrained target $\pi^*(x)$ inherits the score $\nabla \log \pi^*(x)$ from the prior, which typically carries no useful information about the energy landscape. The dominant paradigm for extending nested sampling to high dimensions in general problems has been to employ reflective Hamiltonian dynamics~\citep{skilling_galilean_2019,lemos_improving_2024}. However, these methods have been unable to meaningfully change the scaling of the method and have been shown to be biased in high dimensions~\citep{kroupa_resonances_2025}. Instead of trying to construct an MCMC kernel that can efficiently sample under a hard constraint, we take a different approach and ask if we can soften the hard constraint whilst preserving the properties of the path of monotonically decreasing energy.

The work of \citet{salomone_unbiased_2025} has addressed the consistency and bias of normalizing constant estimation by framing nested sampling as a Sequential Monte Carlo method. We follow the volume accounting from this work, using incremental importance weights to measure the volume ratio between successive levels rather than assigning it from the order-statistic law of the classic nested sampling algorithm~\citep{chopin_properties_2010}. For a fixed schedule, this affords the unbiasedness results central to SMC~\citep{del_moral_feynman-kac_2004}. We nevertheless retain and recommend an inherently adaptive level schedule. This is both natural for the quenched path with its moving support and immensely practical for phase transitions. As in adaptive SMC more generally, this introduces a finite-particle bias~\citep{beskos_convergence_2016}. We separate the measured-volume accounting from the effect of adaptation in \Cref{app:measured-volumes,app:adaptive-levels}.

\subsection{Scalable MCMC kernels}\label{sec:scalable_mcmc}

Setting aside distributions with hard constraints, there has been substantial progress in the robust implementation of scalable Markov chain Monte Carlo (MCMC) methods~\citep{fearnhead_scalable_2025}. By incorporating the gradient of the energy function, $\nabla U(x)$, these methods can suppress the random-walk behaviour of vanilla MCMC. Under standard assumptions, random-walk methods exhibit $\mathcal{O}(D)$ scaling~\citep{roberts_complexity_2016}, whereas Hamiltonian Monte Carlo (HMC)~\citep{duane_hybrid_1987,neal_mcmc_2011} and the Metropolis-adjusted Langevin algorithm (MALA)~\citep{roberts_exponential_1996,roberts_optimal_1998} can achieve $\mathcal{O}(D^{1/4})$ and $\mathcal{O}(D^{1/3})$ scaling, respectively~\citep{roberts_complexity_2016,beskos_optimal_2013}. Incorporating such scalable MCMC kernels into tempered SMC has enabled applications to problems with thousands of dimensions~\citep{buchholz_adaptive_2021}.

This is challenging however for nested sampling, as many recent developments in scalable and efficient MCMC build on these ideas and do not transfer cleanly to hard-constrained targets. Softening the constraint makes innovations in gradient based MCMC kernels such as the Barker proposal MCMC~\citep{livingstone_barker_2022} and isokinetic samplers~\citep{robnik_microcanonical_2023} available to the quenched path. In this work, we use MALA as a proof of principle and develop a recipe for a well-tuned kernel, using the particle ensemble to precondition the proposal and permit efficient application across a variety of problems without hand-tuning. We apply the same kernel within adaptive tempered SMC as a control, providing the key comparison that isolates the effect of the quenched path from that of the MCMC kernel itself.

MALA uses the score of the target density to add a drift term to the standard random-walk Metropolis proposal. The kernel acts on a particle $x$ by proposing a new position $x'$ according to
\begin{equation}
  x' = x + \frac{\epsilon^2}{2}\,M^{-1}\nabla \log \rho(x)
  + \epsilon\,M^{-1/2}\eta,
  \qquad \eta \sim \mathcal{N}(0,I),
\end{equation}
where $\epsilon$ is the step size and $M$ is a positive-definite mass matrix, so the proposal covariance is $\epsilon^2 M^{-1}$. The proposal is then accepted with probability
\begin{equation}
  \alpha(x, x') = \min\Bigl(1, \frac{\rho(x')\,q(x \mid x')}{\rho(x)\,q(x' \mid x)}\Bigr),
\end{equation}
where $q(x' \mid x)$ is the proposal density. The step size $\epsilon$ is typically tuned to achieve an optimal acceptance rate, which is known to be around $0.574$ for MALA in high dimensions~\citep{roberts_optimal_1998}.

\section{Method}
\label{sec:method}

Quenching a configuration in molecular simulation commonly refers to rapidly relaxing it towards an energy minimum, typically performed with fast gradient descent methods~\citep{thompson_lammps_2022}. We present Quenched Ensemble Sampling (\method), a Sequential Monte Carlo sampler that generalises this idea to a population of particles. The population is evolved down a ladder of decreasing energy levels by a combination of reweighting, resampling, and mutation, and the evidence is obtained from the measured level volumes. The resulting sequence of distributions that the particles follow simulates a series of thermodynamic quenches. 

\subsection{The softened level set}\label{sec:softened}

\begin{figure}[t]
\begin{center}
\includegraphics[width=\textwidth]{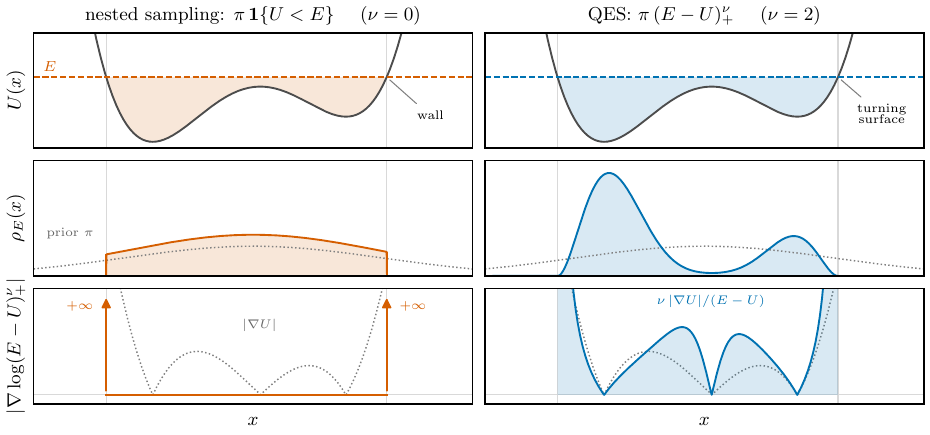}
\end{center}
\caption{Illustration of \method against nested sampling, on a
one-dimensional asymmetric double well (unit Gaussian prior).
\emph{Top:} the energy landscape $U$ with the level $E$ as a waterline. \emph{Middle:} the profile of $\rho_E$ under the two choices of $\nu$.
\emph{Bottom:} the resulting score of $\rho_E$; for $\nu > 0$ it is a
smooth informative gradient throughout the interior of the level set,
vanishing only at stationary points of $U$. The dotted reference is the magnitude of the gradient of $U$.}
\label{fig:sketch}
\end{figure}

At a given energy level $E$, we define a target
\begin{equation}
  \rho_E(x) \;\propto\; \pi(x)\,\bigl(E - U(x)\bigr)_+^{\nu},
  \label{eq:family}
\end{equation}
where $\nu > -1$ is a fixed parameter, and $(\cdot)_+ = \max(\cdot, 0)$. The soft microcanonical factor $(E - U)_+^{\nu}$ is hence a repulsive potential at the level, so that $\rho_E$ vanishes continuously at the boundary of its support $\{U < E\}$. Setting $\nu = 0$ recovers the hard constrained prior at energy $E$, which is the target of nested sampling, and $\nu$ can be set arbitrarily large to increase the repulsive force from the boundary. Pragmatically we use $\nu = 2$ for the mobility and efficiency reasons detailed in~\Cref{app:nu}, and find this choice effective for most applications. The support of $\rho_E$ is still confined to the level set $\{U < E\}$, but its boundary is softened such that the score is defined throughout so can inform MCMC dynamics (\Cref{fig:sketch}).

This family is the conditional of an exact augmentation at fixed temperature. Fix $\nu > -1$ and
consider the joint density on $(x, E)$
\begin{equation}
  p(x, E) \;\propto\; \pi(x)\,(E - U(x))_+^{\nu}\, e^{-E}.
  \label{eq:augmentation}
\end{equation}
Integrating out $E$ with the substitution $s = E - U(x)$ gives
$\int_0^\infty s^\nu e^{-(s + U)}\,\mathrm{d}s = \Gamma(\nu+1)\,e^{-U(x)}$, so
the $x$-marginal of \Cref{eq:augmentation} is exactly the posterior, with
$\Gamma(\nu+1)$ the only constant. Its conditionals are
\begin{equation}
  x \mid E \;\sim\; \rho_E, \qquad
  E \mid x \;\sim\; U(x) + \mathrm{Gamma}(\nu+1, 1).
\end{equation}
Where $E$ is styled as an auxiliary slice variable. The $E$
conditional also informs the meaning of $\nu$, $\nu + 1$ is the mean energy
headroom, in nats, between the level and a particle's own energy. As this is a scale defined in energy (or log-likelihood) space, we see that picking a fixed $\nu$ is justifiable, and this headroom need not necessarily scale with dimension.

Closely related power-law targets arise in microcanonical Monte Carlo
methods~\citep{ray_microcanonical_1991}. In the nested sampling context,
\citet{habeck_nested_2015} introduced auxiliary variable ``demon''
constructions, while \citet{baldock_constant-pressure_2017} developed a related
total-enthalpy HMC method for atomistic systems. These constructions tie the
choice of $\nu$ to the number of degrees of freedom in the augmented system.
QES instead fixes $\nu$ independently of both problem and dimension, and
measures successive level-volume ratios using SMC importance weights rather
than assigning them through nested-sampling order statistics. The inner
mutation admits any kernel that can consume the differentiable log density
in~\Cref{eq:family}, allowing flexible implementation of modern gradient-based
MCMC methods. This makes the soft microcanonical approach directly applicable
to general Bayesian inference problems and, in our experiments, substantially
improves upon classical nested sampling.

\subsection{Partition function from level volumes}\label{sec:partition}

The cumulative density of states defined by~\Cref{eq:dos} gives the level volume
$G(E) = \int_{-\infty}^{E} g(u)\,\mathrm{d}u$. This extends naturally to the
soft microcanonical family of~\Cref{eq:family}, whose level volume is
\begin{equation}\label{eq:softdos}
  G_\nu(E)
  \;=\;
  \mathbb{E}_\pi\bigl[(E-U)_+^\nu\bigr]
  \;=\;
  \int_{-\infty}^{E} g(u)\,(E-u)^\nu\,\mathrm{d}u.
\end{equation}
The same Gamma identity used above expresses the evidence as the
one-dimensional integral
\begin{equation}
  Z
  \;=\;
  \frac{1}{\Gamma(\nu+1)}
  \int_{-\infty}^{\infty} G_\nu(E)\,e^{-E}\,\mathrm{d}E,
  \label{eq:soft-evidence}
\end{equation}
valid for every $\nu>-1$.

Consecutive levels $E'<E$ are related by the pointwise ratio
\begin{equation}
  r(x)
  \;=\;
  \left(
    \frac{(E'-U(x))_+}{(E-U(x))_+}
  \right)^\nu,
  \qquad
  \frac{G_\nu(E')}{G_\nu(E)}
  \;=\;
  \mathbb{E}_{\rho_E}\bigl[r\bigr].
  \label{eq:ratio}
\end{equation}
Given a weighted particle population
$\{(x_i,w_i)\}_{i=1}^N$ targeting $\rho_E$, with
$\sum_i w_i=1$, the incremental ratio is estimated before mutation by
\begin{equation}
  \widehat r_{E\to E'}
  \;=\;
  \sum_{i=1}^N w_i\,r(x_i),
  \qquad
  \widehat G_\nu(E')
  \;=\;
  \widehat G_\nu(E)\,\widehat r_{E\to E'}.
\end{equation}
The same reweighting moves the population to the next target, since
\begin{equation}
  \rho_E(x)\,r(x)
  \;=\;
  \frac{G_\nu(E')}{G_\nu(E)}\,\rho_{E'}(x).
\end{equation}
After a resampling or branching step, an MCMC kernel invariant for $\rho_{E'}$ is
applied to rejuvenate the population and move duplicated particles apart. Repeating
these steps down the level ladder estimates $G_\nu$ through a telescoping
product of importance-weight averages. This is the standard SMC
normalising constant estimator~\citep{doucet_introduction_2001}, applied
here to softened level-set volumes as the path to the marginal
likelihood.

\subsection{Algorithm}\label{sec:algorithm}

With the definitions of the previous sections, we can now compose the complete
algorithm. We assume that the reference distribution $\pi$ is easy to sample
and that the energy $U$ is differentiable. With a target population of $N$
particles, the algorithm can be initialised by drawing $N$ samples from $\pi$
and setting the initial level $E_0$ to the maximum energy in that population.
Although this is a valid starting point, in practice we find it more robust to
draw a larger anchor population of, say, $10N$ samples and choose $E_0$ by
bisection until the importance weights $(E_0-U)_+^\nu$ have effective sample
size (ESS) $N$. The empirical mean of these weights initializes $G_\nu(E_0)$,
and their normalized values are resampled to form the $N$-particle working
population (\Cref{app:initialization}). This protects the initial importance
step from extreme weights in the tail of the prior-energy distribution.

Given $N$ particles approximately distributed as $\rho_{E_k}$, we can apply the algorithm described in \Cref{alg:quenched_sampling} to transition to the next level $E_{k+1} < E_k$. The algorithm consists of three main steps: (1) dissecting the current population to choose the next energy level, (2) reweighting and branching the population to account for the new level, and (3) mutating the particles with an MCMC kernel invariant to $\rho_{E_{k+1}}$. The next level is found by bisection until the prospective reweighted population reaches a prescribed effective sample size. Here $K_E$ denotes the complete inner MCMC mutation block: its construction and number of internal steps are unrestricted, provided that it leaves $\rho_E$ invariant. The process is repeated until the contribution to the marginal likelihood from the active particles becomes negligible compared to the accumulated marginal likelihood, at which point we terminate and compute the final estimate of $Z$ using quadrature on the ladder of levels.

\begin{algorithm}[htb]
  \caption{One quenched level transition}
  \label{alg:quenched_sampling}
  \begin{algorithmic}[1]
    \Require Weighted ensemble $\{(x_j,w_j)\}_{j=1}^N$ targeting
      $\rho_{E_k}$, with $\sum_j w_j=1$
    \Require Current level volume $\widehat G_{\nu,k}$ and an MCMC
      kernel $K_E$ that leaves $\rho_E$ invariant
    \State \textbf{Dissect.} Choose $E_{k+1}<E_k$ by bisection until
      the ESS of $\{w_jr(x_j)\}_{j=1}^N$ reaches its target.
    \State \textbf{Measure.} At the pre-mutation positions, compute
      $\widehat r_k\gets\sum_j w_jr(x_j)$ and
      $\widehat G_{\nu,k+1}\gets\widehat G_{\nu,k}\widehat r_k$.
    \State \textbf{Reweight.} Set
      $w_j\gets w_jr(x_j)/\widehat r_k$ for every particle.
    \State \textbf{Branch.} Replace each zero-weight particle by a copy of
      a survivor drawn by weight; assign half the donor weight to each copy.
    \State \textbf{Resample.} When required, based on a second ESS threshold, resample the ensemble and reset
      all $w_j\gets1/N$.
    \State \textbf{Mutate.} Apply $K_{E_{k+1}}$ independently to every
      particle.
    \Ensure $\{(x_j,w_j)\}_{j=1}^N$, $E_{k+1}$, and
      $\widehat G_{\nu,k+1}$
  \end{algorithmic}
\end{algorithm}

\paragraph{Choice of mutation kernel.} The estimator requires an invariant mutation kernel, where there are a vast array of choices. We employ Metropolis-adjusted
Langevin dynamics as detailed in~\Cref{sec:scalable_mcmc}, with a kernel pre-tuning recipe defined in this section. This proves to be a relatively robust choice, and the automated tuning we implement is effective for a variety of problems covered in \Cref{sec:experiments}, this also defines a strong baseline for tempered SMC which we provide as well in the code implementation. Following the pretuning pattern of \citet{buchholz_adaptive_2021}, we use the accumulated gradient information from the previous level to construct a diagonal preconditioner. Writing this score $s^{(i)} = \nabla \log \rho\bigl(x^{(i)}\bigr)$ for the score at particle $i$ of the ensemble, we set $M^{-1} = \operatorname{diag}(\sigma_1^2, \ldots, \sigma_D^2)$ with
\begin{equation}
  \sigma_j^{-1} \;=\; \operatorname*{median}_i \bigl\lVert s^{(i)} \bigr\rVert \,
  \biggl\langle \frac{s_j^{2}}{\lVert s \rVert^{2}} \biggr\rangle^{1/2},
  \label{eq:metric}
\end{equation}
where $\langle \cdot \rangle$ averages over mutation steps and the weighted ensemble, and the scale is the ensemble median of each walker's step-averaged norm. This is a natural choice: the anisotropy $\langle s_j^2/\lVert s\rVert^2\rangle$ is the direction normalised Fisher information metric of the target, frozen across the mutation chain~\citep{girolami_riemann_2011}. Rather than using the mean of the score norm as a scale we use the median since the additional boundary proximity term $(E-U)^{-1}$ that appears in this particular score can diverge for particles near the boundary, and the median is robust to this. The step size $\epsilon$ is adapted across levels by constant-gain stochastic approximation~\citep{robbins_stochastic_1951}, driving the acceptance rate observed at the previous level towards $0.574$, the gain is held constant rather than decayed, since $E$ moves at every level and the optimal step size drifts with it.

Two properties of \Cref{eq:metric} are important for embedding in a particle method. First, the metric is read from the score rather than from the spread of the particles. The particle cloud covariance, the nested sampling default~\citep{yallup_nested_2026-1}, estimates the extent of the ensemble. On a multimodal target that extent is the separation between modes rather than the width of any one of them, so the metric degrades on the type of problem that particle methods are commonly invoked on. Using the averaged local score of~\Cref{eq:metric} composes only the local property of the target, so is well defined even on massively multimodal targets such as Bayesian neural networks. Second, neither $M$ nor $\epsilon$ changes while the mutation kernel is applied to the level set. Both are constants of a level, set from the ensemble at the level above, maintaining the required invariance for an unbiased estimator. The measured acceptance stays in the range $0.53$--$0.57$ throughout \Cref{sec:experiments}.

\paragraph{The adaptive schedule.} Choosing $E_{k+1}$ from the current
population makes the schedule adaptive, and adaptive schedules cost exact
unbiasedness. The estimator stays consistent, but $\mathbb{E}[\hat{Z}] = Z$ no
longer holds at finite $N$. This is a property of adaptive SMC generally, and of nested sampling in particular, whose energy threshold is
likewise read off the current particles---\citet{salomone_unbiased_2025} show that the
fixed-threshold form of nested sampling is unbiased while the adaptive form is
only consistent. The nature of the moving maximum $E$ is that progressing to $E_{k+1}$ pushes the target fraction of the population out of the support of $\rho_{E_{k+1}}$, this form of progression naturally benefits from an adaptive schedule. If too large a jump is taken in tempering the population collapses to a single particle, in the case of nested sampling it could cause all particles to be outside the support of the next target, and the algorithm would fail. Hence, initially running with an adaptive schedule is a practical necessity, and we find that it does not introduce a significant bias in practice; its finite-particle consequence is detailed in \Cref{app:adaptive-levels}. Due to the length of the ladder of energy levels that naturally emerges on the quenched path, we hold a second ESS target for resampling the full population weights, which we implement using systematic resampling. As detailed in \Cref{app:resample}, this allows many energy levels to be crossed without resampling, which at the cadence of the quenched path is often unnecessary and can be detrimental to population diversity.

\paragraph{Termination.} 

Unlike SMC, where the invariant target is the canonical ensemble, quenched sampling would theoretically continue quenching until the energy of each particle is at the global minimum. One can either truncate at a chosen energy level, or use the standard criterion of \citet{skilling_nested_2006}, estimating the remaining integral mass from $G_\nu(E)\,e^{-U_{\min}}$,
where $U_{\min}$ is the current lowest particle energy, and stopping when this estimate falls below $e^{\dlogz}$ times the accumulated integral. We adopt $\dlogz=-3$ as a default, which is standard for typical Bayesian posterior targets~\citep{handley_polychord_2015}. For targets with a first-order transition, the threshold must be deepened so that termination occurs beyond the transition rather than in the broad phase, as detailed in \Cref{app:config}.
Finally \Cref{eq:soft-evidence} can be evaluated by quadrature on
the ladder of quenched transitions. Equally weighted posterior draws can be obtained from the augmentation
in~\Cref{eq:augmentation} by sampling a level with probability proportional to
$G_\nu(E)e^{-E}\mathrm{d}E$, then subsequently sampling a particle from that level's cloud. The full
weighted-pool construction is detailed in~\Cref{app:pooling}.

\section{Experiments}
\label{sec:experiments}

We demonstrate the method on three sets of experiments.
\begin{enumerate}[label=(\roman*)]
  \item Analytic targets, including a Gaussian with no transition and a
    spike--slab mixture with a first-order transition, to test scalability and
    accuracy against exact results (\Cref{sec:analytic}).
  \item Bayesian neural network inference and model comparison on UCI
    classification benchmarks (\Cref{sec:bnn}).
  \item A lattice $\phi^6$ field theory with a first-order transition, to test
    traversal of a discontinuous phase transition at scale in a physical model
    (\Cref{sec:lattice}).
\end{enumerate}

Across the examples we primarily compare against tempered SMC using the same
inner kernel and tuning. For the BNN tasks we also compare against scalable
MCMC methods, including NUTS~\citep{hoffman_no-u-turn_2014} and
MCLMC~\citep{robnik_microcanonical_2023}, and on the analytic targets we
include a gradient-free nested sampling baseline~\citep{yallup_nested_2026}.
Experimental configurations are detailed in~\Cref{app:config}. All experiments
are implemented in \texttt{JAX}~\citep{bradbury_jax_2018} using
\texttt{blackjax}~\citep{cabezas_blackjax_2024} and run on a
\citet{google_cloud_tpu_2024} TPU v6e (Trillium) accelerator. All sampling uses single precision
\texttt{float32} numerics, enabling efficient computation on a variety of accelerator backends.

\subsection{Analytic targets}
\label{sec:analytic}

We summarize QES against tempered SMC using the same inner MCMC kernel, and
against classic nested sampling using slice sampling, in~\Cref{tab:analytic}.
All three methods use $N=1000$ particles or live points. QES and tempered SMC
choose their next level at a target ESS of $0.95$. Their MALA mutation budget
is $\max(16,\sqrt{D})$ steps per level, rounded up to the next power of two;
nested sampling instead uses $2D$ slice steps per replacement. In QES, full
resampling is triggered separately when the carried-weight ESS falls below
$0.5N$, as motivated in~\Cref{app:resample}.

The Gaussian target combines a unit Gaussian prior with a narrower likelihood,
$\sigma=0.5$, and tests scaling over $D\in\{10,100,1000,10000\}$. QES remains
accurate well beyond the range reached by the nested sampling baseline, although
a small sub-nat residual bias emerges at $D=10^4$. We attribute this to the increased length of the QES path along with the accumulated finite-$N$ effects, as detailed in~\Cref{app:accounting}. Even at $D=10^2$, the baseline nested sampling implementation shows some emerging bias, which is due primarily to imperfect mixing of the slice kernel at this scale and fixed mutation budget. Tempered SMC is more accurate at $D=10^4$, as expected on this unimodal target.
At fixed target ESS, the number of quenched
levels grows linearly with dimension, compared with $\sqrt{D}$ for tempering,
so QES requires more gradient evaluations. However, the number of quenched
levels carrying appreciable posterior weight grows as $\sqrt{D}$, while it
remains roughly constant for tempering. From $D=100$ onward on this benchmark, the
larger QES pool outweighs the longer path and gives a higher pooled Kish ESS
per gradient evaluation (\Cref{app:typical}).

Quenching distinguishes itself from tempering on the spike--slab benchmark: a
diffuse Gaussian slab and a narrow spike at the origin, scaled so that the spike
carries $90\%$ of the posterior mass. This provides a controlled first-order
transition. Reaching the spike requires a deeper termination threshold for QES
and nested sampling. Across the tested dimensions and ESS schedules, tempered
SMC remains approximately $2.3$ nats low, whereas the QES estimates are
consistent with the analytic value within the seed variation. The large
nominal SMC Kish ESS on these runs measures weight concentration within the
sampled phase and does not certify convergence of the path.

Finally, we test a strongly anisotropic bimodal mixture whose coordinate
standard deviations range from $10^{-2}$ to $1$. The modes are not traversed by
a local chain within the tested mutation budget, so the particle population
must preserve both. Because losing one symmetric mode leaves the marginal
likelihood unchanged, we instead report mode occupancy, whose exact value is
$0.5$. Both QES and tempered SMC recover the balance on average
(\Cref{app:resample}), and their
pooled Kish ESS per gradient becomes comparable as the dimension grows.

\begin{table}[t]
\caption{The analytic test problems (target definitions in
\Cref{app:targets}): evidence bias against the exact $\log Z$ and
pooled Kish ESS per $10^6$ gradient evaluations. Gaussian evidence errors use
ten seeds; the ESS values and remaining results use three.
The bimodal test problem tests against \emph{mode occupancy}, the
diagnostic that detects mode loss. The ESS convention is one Kish calculation
after pooling the full per-particle weight vectors from every rung, as defined in
\Cref{app:pooling}, divided by that run's own gradient count. NS uses slice sampling and counts likelihood evaluations as opposed to gradients, dashes mark cells not run due to computational cost. Tempering is consistent with the analytic value where
there is no transition and $2.3$ nats low wherever there is one,
while matching \method on mode balance. \method starts to exhibit a residual bias on the Gaussian at $D=10^4$.}
\label{tab:analytic}
\begin{center}
\begin{tabular}{@{}llrrrrrr@{}}
\toprule
 & & \multicolumn{3}{c}{$\Delta \log Z$} & \multicolumn{3}{c}{ESS per $10^6$ gradients} \\
\cmidrule(lr){3-5}\cmidrule(l){6-8}
target & $D$ & \method & SMC & NS & \method & SMC & NS \\
\midrule
Gaussian & $10$ & $-0.02$\pmerr{0.04} & $+0.01$\pmerr{0.03} & $+0.01$\pmerr{0.08} & $21519$ & $26507$ & $126$ \\
 & $100$ & $-0.02$\pmerr{0.08} & $+0.02$\pmerr{0.05} & $+0.66$\pmerr{0.07} & $13384$ & $7933$ & $2.47$ \\
 & $1000$ & $-0.21$\pmerr{0.36} & $-0.02$\pmerr{0.12} & --- & $2972$ & $1228$ & --- \\
 & $10000$ & $-0.89$\pmerr{0.71} & $-0.12$\pmerr{0.18} & --- & $266$ & $101$ & --- \\
\midrule
spike--slab & $10$ & $-0.01$\pmerr{0.03} & $-2.29$\pmerr{0.00} & $+0.09$\pmerr{0.09} & $6564$ & $48393$ & $19$ \\
 & $50$ & $+0.16$\pmerr{0.21} & $-2.30$\pmerr{0.00} & $+0.19$\pmerr{0.13} & $3707$ & $26233$ & $0.94$ \\
 & $200$ & $+0.04$\pmerr{0.32} & $-2.32$\pmerr{0.01} & --- & $2698$ & $13168$ & --- \\
 & $500$ & $+0.22$\pmerr{0.12} & $-2.32$\pmerr{0.05} & --- & $702$ & $3957$ & --- \\
\midrule
 & & \multicolumn{3}{c}{mode occupancy (true $0.5$)} & \multicolumn{3}{c}{ESS per $10^6$ gradients} \\
\cmidrule(lr){3-5}\cmidrule(l){6-8}
 & $D$ & \method & SMC & NS & \method & SMC & NS \\
\midrule
bimodal & $10$ & $0.49$\pmerr{0.24} & $0.51$\pmerr{0.02} & $0.51$\pmerr{0.22} & $4494$ & $7080$ & $7.77$ \\
 & $50$ & $0.60$\pmerr{0.19} & $0.49$\pmerr{0.04} & $0.61$\pmerr{0.16} & $2679$ & $2916$ & $0.46$ \\
 & $100$ & $0.46$\pmerr{0.13} & $0.45$\pmerr{0.05} & --- & $2044$ & $1984$ & --- \\
 & $200$ & $0.63$\pmerr{0.25} & $0.52$\pmerr{0.20} & --- & $1479$ & $1448$ & --- \\
\bottomrule
\end{tabular}

\end{center}
\end{table}

\begin{figure}[t]
\begin{center}
\includegraphics[width=0.48\textwidth]{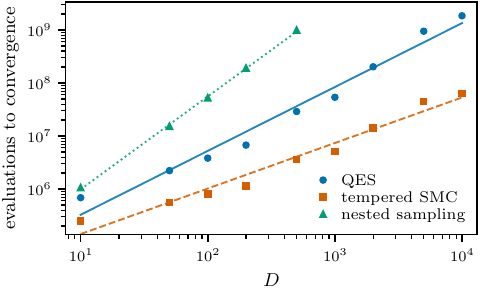}
\hfill
\includegraphics[width=0.48\textwidth]{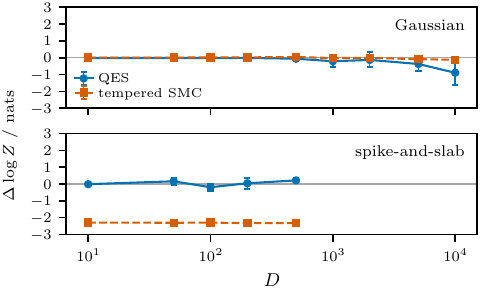}
\end{center}
\caption{Cost and accuracy order the three methods oppositely. \emph{Left:}
evaluations to convergence against dimension on the Gaussian, each method run
to its own criterion at $N = 1000$, with a fitted power law
$\text{evals} = C D^{\alpha}$. The cost exponents are $\alpha = 0.86$ for tempered SMC, $1.21$ for \method and
$1.75$ for nested sampling. \emph{Right:} evidence bias on a shared
symmetric axis, Gaussian and spike--slab. Nested sampling counts likelihood evaluations, the other two count gradients,
so cost is not exactly comparable.}
\label{fig:scaling}
\end{figure}

\begin{figure}[t]
\begin{center}
\includegraphics[width=\textwidth]{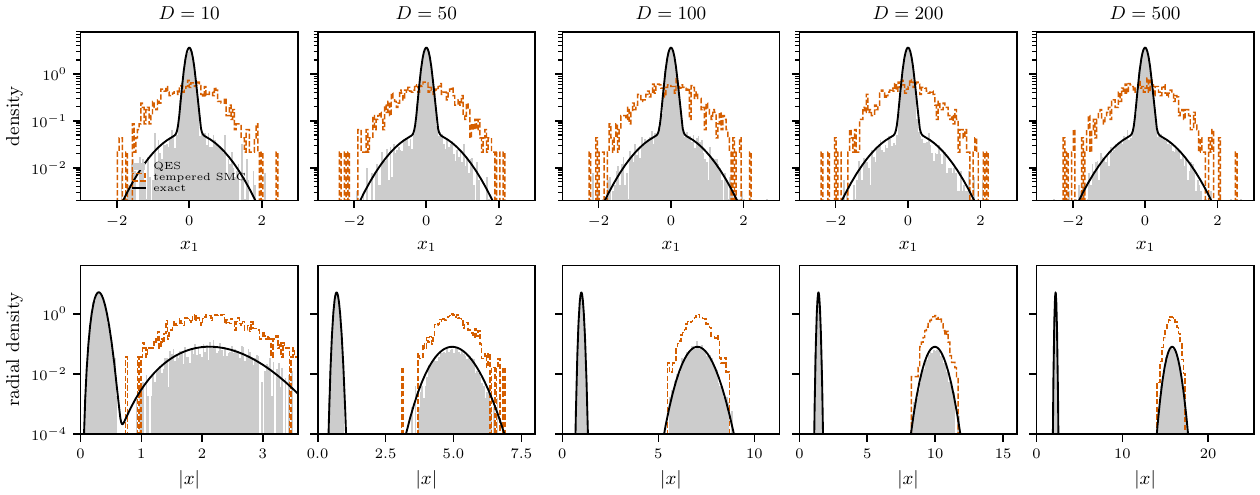}
\end{center}
\caption{Posterior reconstruction on the spike--slab target against closed
form. \emph{Top:} marginal in the first coordinate on a logarithmic density
axis, so the slab---which carries $10\%$ of the evidence but is ten times
wider---is visible beside the spike. \emph{Bottom:} radial density, where the
two phases are disjoint shells at $|x| = \tau_j\sqrt{D}$ and the coexistence
gap is explicit, empty to within sampling noise over four decades and widening
with dimension. The spike is slightly under-weighted at the largest two
dimensions, although the corresponding evidence estimates remain consistent
with the analytic values within the seed variation.}
\label{fig:marginal}
\end{figure}

\subsection{Bayesian neural networks}
\label{sec:bnn}

A Bayesian treatment of neural networks places a prior on the weights and biases of a network and attempts to sample from the posterior distribution of these parameters given a training dataset~\citep{izmailov_what_2021}. As well as being highly multimodal, high dimensional and having challenging posterior geometry to sample from, it is understood that the training of networks can exhibit phase transition properties~\citep{montanari_interpolation_2022,power_grokking_2022}. We consider this task a demonstration of three facets of the method: the ability to competitively scale to (small) neural network scale problems with other state-of-the-art samplers, the ability to return a marginal likelihood estimate for model comparison, and the ability to scan for phase transitions in the posterior landscape.

We take benchmark neural network fitting tasks from
\citet{sommer_microcanonical_2024}: five classification datasets from the UCI repository~\citep{kelly_uci_nodate}. Each task is fit
with a fully connected $2\times16$ $\tanh$ network, giving between $528$ and
$1280$ parameters depending on the input data dimension. We place independent
standard normal priors on the network parameters and scale each layer's weight
contribution by $1/\sqrt{d_{\mathrm{in}}}$, where $d_{\mathrm{in}}$ is the number of input connections to each neuron, this ensures effective weights have variance
$1/d_{\mathrm{in}}$. We use a fixed 90--10 train--test split, fit a full batch, and report
efficiency as ESS per million gradient evaluations and predictive accuracy as
the posterior-predictive mixture NLL as defined in \Cref{app:targets}.

QES and tempered SMC are matched using the same population of $1000$
particles, the same inner kernel and preconditioner, $96$ inner MCMC steps per
level, a target ESS of $0.9$, and a convergence criterion of
$\dlogz=-3$ for QES. We additionally compare against ensembles of NUTS
chains~\citep{hoffman_no-u-turn_2014} and MCLMC
chains~\citep{robnik_microcanonical_2023}, using standard window adaptation for
NUTS and the MCLMC tuning procedure of
\citet{sommer_microcanonical_2024} (with the deep ensemble initialisation omitted). We use $1000$ chains to match
the particle methods' parallel ensemble size, and normalize efficiency by each
method's actual number of gradient evaluations, including warmup and tuning.
The full chain budgets and ESS conventions are given in the table caption and
\Cref{app:config}, where we note that strict comparison between particle methods and chains is difficult due to difference in implementation. As summarized in~\Cref{tab:clf}, all methods
give very similar predictive NLL, with MCLMC trailing slightly on two of the
five tasks, while NUTS is the most efficient under the tested budgets. At comparable predictive accuracy, QES returns
between $1.4$ and $2.0$ times the pooled Kish ESS per gradient of tempered SMC
across the five tasks, while successfully exploring these BNN posteriors from
prior initialization.

Secondly we use the derived marginal likelihood estimates to compare the choice of activation function on the same $2\times16$ networks. In this setting we introduce an additional approximate baseline, we construct a Laplace
approximation from a MAP estimate found by Adam~\citep{kingma_adam_2015}, using
a generalised Gauss--Newton curvature~\citep{duffield_scalable_2024}. We report the resulting Bayes factors, as well as the implications for which fit is preferred in~\Cref{tab:clfbf}. We find that a Laplace approximation gives a poor estimate and picks inconsistent activations, whereas both particle methods are in close agreement on all datasets. This agreement validates that it is possible to use the marginal likelihood estimates from QES for model comparison in Bayesian neural networks at this scale.

\begin{figure}[htb]
  \begin{center}
  \includegraphics[width=\linewidth]{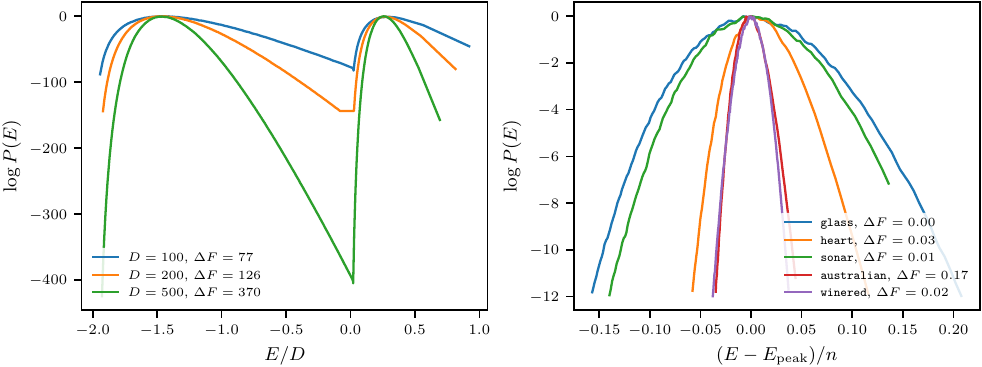}
  \caption{Scan for phase transitions in the level marginal
  $P(E) \propto G_\nu(E)\,e^{-E}$, the integrand of
  \Cref{eq:soft-evidence}, read from a single ladder per curve. \emph{Left:} the
  spike--slab posteriors of \Cref{sec:analytic} are bimodal with a
  free-energy barrier $\Delta F$ that grows extensively with dimension ---
  a first-order transition. \emph{Right:} the UCI classification posteriors at
  comparable dimension show no evidence of first-order phase structure at the
  resolution probed, across a deep energy range. The energy axis is made intensive, by dimensionality for the spike--slab and by training set size for the networks, with the network profiles additionally centred on the posterior modal energy $E_\text{peak}$ so that they occupy a similar range.}
  \label{fig:barrier}
  \end{center}

\end{figure}

Lastly, we search for phase transitions in the full batch likelihood. For this we extend the convergence criterion from $\dlogz=-3$ to the deeper values used in the spike--slab experiments, and reconstruct the level marginal $\log P(E)$---the integrand of~\Cref{eq:soft-evidence}, read directly off the quenched ladder. In terms of the free-energy profile $F(E) = -\log P(E)$, a first-order transition appears as two peaks in $P(E)$ separated by a valley of suppressed interfacial states~\citep{lee_finite-size_1991}; the barrier $\Delta F$ is the depth of that valley below the lower peak, in nats. By \Cref{eq:augmentation}, $E = U(x) + s$ with independent slack $s \sim \Gamma(\nu+1,1)$, so $P(E)$ is the $\beta=1$ posterior energy distribution under a fixed smoothing of width $\sqrt{\nu+1}$ nats---invisible to extensive first-order structure, but blurring features narrower than a few nats. Bimodality in $P(E)$ therefore identifies a phase transition, and a barrier growing with system size, as on the left, identifies it as first order. \Cref{fig:barrier} shows this profile for the five classification tasks against the spike--slab target of~\Cref{sec:analytic} at comparable dimensionality. Both panels use intensive energy axes so that curves of different size share a scale: the spike--slab energy is a sum over the $D$ dimensions and is plotted per dimension, while the network energy is a sum over the $n$ training points and is plotted per datapoint, centred on its peak, since the raw spans differ by nearly an order of magnitude across datasets. The network posteriors show no evidence for a phase transition on these architectures and tasks. Studying these phenomena empirically in finite-width networks on real data remains an open question, and quenched sampling provides a unique tool for this investigation.

\begin{table}[t]
\caption{Posterior comparison on UCI classification tasks, each metric is averaged across three seeds.
Performance is measured by the posterior-predictive NLL on a held out test set, and the efficiency is measured by the ESS per million gradient evaluations, including warmup and tuning. For each particle run, one Kish calculation pools the per-particle weights from every rung of the
ladder. The chain methods get the rank-normalised bulk ESS of
\citet{vehtari2021rank} over all $1000$ chains jointly, which is not strictly comparable to the pooled Kish ESS but can be used as a qualitative comparison. Corresponding model comparison results are given in~\Cref{tab:clfbf}.}
\label{tab:clf}
\begin{center}
\resizebox{\textwidth}{!}{\begin{tabular}{@{}lrrrrrrrrr@{}}
\toprule
 & & \multicolumn{4}{c}{test NLL} & \multicolumn{4}{c}{ESS per $10^6$ gradients} \\
\cmidrule(lr){3-6}\cmidrule(l){7-10}
 & $D$ & \method & SMC & NUTS & MCLMC & \method & SMC & NUTS & MCLMC \\
\midrule
sonar & $1280$ & $0.458$\pmerr{0.002} & $0.456$\pmerr{0.001} & $0.457$\pmerr{0.003} & $0.472$\pmerr{0.002} & $3184$\pmerr{81} & $2195$\pmerr{152} & $8996$\pmerr{42} & $64$\pmerr{1} \\
glass & $528$ & $0.990$\pmerr{0.002} & $0.991$\pmerr{0.002} & $0.991$\pmerr{0.002} & $1.002$\pmerr{0.001} & $2367$\pmerr{19} & $1362$\pmerr{84} & $11631$\pmerr{38} & $1837$\pmerr{500} \\
heart & $528$ & $0.464$\pmerr{0.000} & $0.465$\pmerr{0.001} & $0.463$\pmerr{0.000} & $0.463$\pmerr{0.001} & $3017$\pmerr{23} & $1836$\pmerr{45} & $10556$\pmerr{68} & $442$\pmerr{23} \\
australian & $544$ & $0.319$\pmerr{0.000} & $0.319$\pmerr{0.001} & $0.319$\pmerr{0.001} & $0.319$\pmerr{0.001} & $2270$\pmerr{8} & $1422$\pmerr{34} & $10804$\pmerr{18} & $1296$\pmerr{138} \\
wine & $560$ & $0.909$\pmerr{0.000} & $0.909$\pmerr{0.001} & $0.909$\pmerr{0.001} & $0.911$\pmerr{0.000} & $1189$\pmerr{10} & $582$\pmerr{32} & $4375$\pmerr{17} & $208$\pmerr{14} \\
\bottomrule
\end{tabular}
}
\end{center}
\end{table}

\begin{table}[t]
\caption{Model comparison on the UCI classification tasks, fixed network dimensions comparing $\tanh$ against ReLU activations. 
$\log \mathrm{BF} = \log \hat{Z}(\tanh) - \log \hat{Z}(\mathrm{ReLU})$ is computed over three repeated seeds and used to compare the two activations where \colorbox{selbg}{shading} marks
the selected model. \method\ and tempered SMC
agree on every dataset and their Bayes factors agree to within $0.6$ nats;
every factor is resolved (the smallest lies $3.5$ seed standard deviations
from zero) while on \texttt{glass}, \texttt{sonar} and \texttt{winered} the
test NLL weakly favours the activation the evidence rejects. The Laplace--GGN
approximation reports factors two to four times larger and selects the
opposite activation on some tasks.}
\label{tab:clfbf}
\begin{center}
\resizebox{\textwidth}{!}{\begin{tabular}{@{}llrrrrrrr@{}}
\toprule
 & & & \multicolumn{3}{c}{test NLL} & \multicolumn{3}{c}{$\log \mathrm{BF}$} \\
\cmidrule(lr){4-6}\cmidrule(l){7-9}
 & & $D$ & \method & SMC & Laplace & \method & SMC & Laplace \\
\midrule
glass & $\tanh$ & $528$ & $0.990$\pmerr{0.002} & $0.991$\pmerr{0.002} & \cellcolor{selbg}$1.231$\pmerr{0.001} & \multirow{2}{*}{$-6.0$\pmerr{0.46}} & \multirow{2}{*}{$-6.2$\pmerr{0.11}} & \multirow{2}{*}{$+2.3$\pmerr{0.39}} \\
 & ReLU &  & \cellcolor{selbg}$0.993$\pmerr{0.005} & \cellcolor{selbg}$0.993$\pmerr{0.001} & $1.181$\pmerr{0.024} &  &  &  \\
\addlinespace
heart & $\tanh$ & $528$ & \cellcolor{selbg}$0.464$\pmerr{0.000} & \cellcolor{selbg}$0.465$\pmerr{0.001} & \cellcolor{selbg}$0.503$\pmerr{0.002} & \multirow{2}{*}{$+7.7$\pmerr{0.46}} & \multirow{2}{*}{$+7.7$\pmerr{0.03}} & \multirow{2}{*}{$+19.6$\pmerr{0.49}} \\
 & ReLU &  & $0.465$\pmerr{0.001} & $0.467$\pmerr{0.002} & $0.495$\pmerr{0.004} &  &  &  \\
\addlinespace
sonar & $\tanh$ & $1280$ & \cellcolor{selbg}$0.458$\pmerr{0.002} & \cellcolor{selbg}$0.456$\pmerr{0.001} & \cellcolor{selbg}$0.534$\pmerr{0.007} & \multirow{2}{*}{$+5.3$\pmerr{0.11}} & \multirow{2}{*}{$+5.2$\pmerr{0.06}} & \multirow{2}{*}{$+15.7$\pmerr{0.45}} \\
 & ReLU &  & $0.444$\pmerr{0.002} & $0.444$\pmerr{0.001} & $0.505$\pmerr{0.004} &  &  &  \\
\addlinespace
australian & $\tanh$ & $544$ & \cellcolor{selbg}$0.319$\pmerr{0.000} & \cellcolor{selbg}$0.319$\pmerr{0.001} & \cellcolor{selbg}$0.428$\pmerr{0.001} & \multirow{2}{*}{$+10.5$\pmerr{0.22}} & \multirow{2}{*}{$+10.6$\pmerr{0.05}} & \multirow{2}{*}{$+36.9$\pmerr{1.55}} \\
 & ReLU &  & $0.324$\pmerr{0.001} & $0.324$\pmerr{0.001} & $0.448$\pmerr{0.027} &  &  &  \\
\addlinespace
winered & $\tanh$ & $560$ & $0.909$\pmerr{0.000} & $0.909$\pmerr{0.001} & \cellcolor{selbg}$1.005$\pmerr{0.003} & \multirow{2}{*}{$-37.3$\pmerr{0.10}} & \multirow{2}{*}{$-37.9$\pmerr{0.16}} & \multirow{2}{*}{$+15.9$\pmerr{4.50}} \\
 & ReLU &  & \cellcolor{selbg}$0.930$\pmerr{0.001} & \cellcolor{selbg}$0.929$\pmerr{0.000} & $0.985$\pmerr{0.007} &  &  &  \\
\bottomrule
\end{tabular}
}
\end{center}
\end{table}

\subsection{Lattice \texorpdfstring{$\phi^6$}{phi^6}: a first-order transition}
\label{sec:lattice}

Finally, we consider the lattice $\phi^6$ model, a two-dimensional scalar field
theory with a first-order transition on a continuous state
space~\citep{makhankov_6_1990,sanati_half-kink_1999}. The field is defined on a
periodic $L\times L$ lattice, with $D=L^2$. We use the exactly samplable free
field at reference mass $m_0^2$ as the prior $\pi_0(\phi)$, and define the
remaining interaction action---the energy function passed to the sampler---as
\begin{equation}
U(\phi) \;=\; \sum_x \Big[ \tfrac{1}{2}\big(\mu^2 - m_0^2\big)\phi_x^2
   \;+\; \lambda\,\phi_x^4 \;+\; \eta\,\phi_x^6 \Big],
\label{eq:phi6}
\end{equation}
where $x$ runs over the lattice sites, so the posterior is $\propto\pi_0(\phi)e^{-U(\phi)}$. With $\lambda<0$
and $\eta>0$, the on-site potential has a pair of
minima that exchange stability with the one at the origin
\emph{discontinuously}, giving a first-order transition rather than the
second-order one of ordinary $\phi^4$. Mean field puts coexistence at
$\mu^2_\star = \lambda^2/2\eta$; at $L = 40$, $\lambda = -1$, $\eta = 0.2$,
$m_0^2 = 1$ the located coexistence point is $\mu^2 = 2.28347$
(\Cref{app:targets}), and the uniform-field barrier separating the phases is
$0.463$ per site, i.e.\ $741$ nats.

We report two order parameters: $m=V^{-1}\sum_x\phi_x$, which identifies the
$Z_2$ parity symmetry sector, and $q=V^{-1}\sum_x\phi_x^2$, which distinguishes the phases,
with $V=L^2$. We classify configurations using the mean-field cut
$q_{\rm cut}=1.25$; at coexistence the two phases carry equal probability. Both
tempering and quenching use $8000$ particles, $64$ inner MCMC steps, and a
target ESS of $0.95$.

\begin{table}[H]
\caption{Lattice $\phi^6$ at coexistence, $L = 40$ ($D = 1600$),
$\mu^2 \approx 2.28$. Both methods use $N = 8000$ particles, $64$ inner steps and the
same preconditioned MALA kernel; the adaptive runs target an ESS of $0.95$.
SMC is shown at two further adaptive schedules and at $10{,}000$ equally spaced
temperatures. Results are averaged over 3 seeds. The QES seeds agree
to $0.01$ nats in $\log \hat{Z}$, and is the only method to resolve both ordered and disordered phases.}
\label{tab:phi6}
\begin{center}
\small
\begin{tabular}{@{}llrrrrr@{}}
\toprule
 & schedule & levels & $\log \hat{Z}$ & ordered share & $\langle |m| \rangle$ & $\langle q \rangle$ \\
\midrule
\method      & $0.95$  & $11502$\pmerr{372} & $-53.76$\pmerr{0.01} & $0.543$\pmerr{0.009} & $0.733$\pmerr{0.011} & $1.112$\pmerr{0.015} \\
\addlinespace
tempered SMC & $0.95$  & $20$    & $-54.54$ & $0.000$ & $0.027$ & $0.184$ \\
             & $0.99$  & $44$    & $-54.54$ & $0.000$ & $0.028$ & $0.184$ \\
             & $0.999$ & $137$   & $-54.54$ & $0.000$ & $0.028$ & $0.184$ \\
             & fixed & $10{,}000$ & $-54.54$ & $0.000$ & $0.028$ & $0.184$ \\
\bottomrule
\end{tabular}
\end{center}
\end{table}

The QES ladder resolves both phases, with an ordered share of
$0.543 \pm 0.009$ over three seeds against the coexistence value of $0.5$. Tempered SMC does not enter the ordered phase
in any tested run, even when its target ESS is increased to $0.999$, and a long
preconditioned HMC chain (window-adapted NUTS, \Cref{app:config}) remains in
the disordered phase. Among the methods
tested, only QES resolves both phases. Doing so requires a termination threshold
deep enough to span the energy barrier and therefore carries a substantial
computational cost. Aside from tempering, a popular approach to this type of problem is to try and flatten the density of states and sample from $\rho(x) \propto \pi(x)/g(U(x))$, adaptively approaching this by constructing a flat binning in energy~\citep{wang_efficient_2001}. Efficient implementation of this approach is challenging in discrete systems~\citep{dayal_performance_2004}, and in continuous systems such as the lattice $\phi^6$ model constructing the optimal binning is non-trivial. We leave a comparison to this approach to future work, however we note that classic nested sampling resembles an optimal choice of binning in energy, discovered on the fly~\citep{partay_efficient_2010}, and QES inherits this property.

\begin{figure}[t]
\begin{center}
\includegraphics[width=\textwidth]{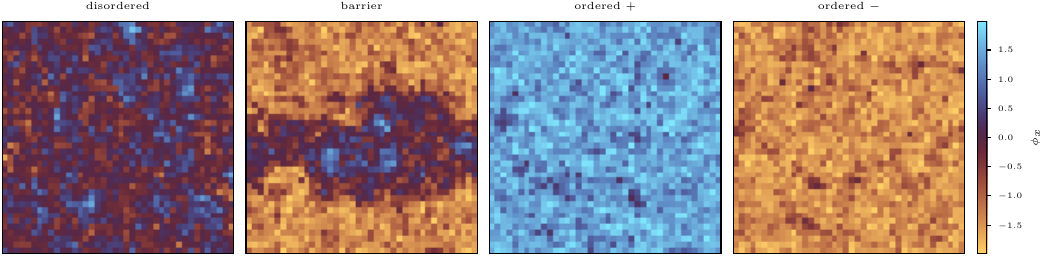}\\[0.35em]
\includegraphics[width=0.85\textwidth]{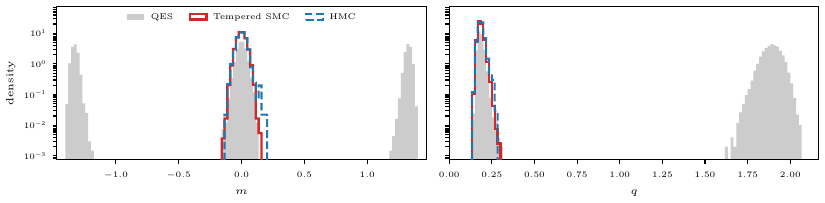}
\end{center}
\caption{Lattice $\phi^6$ at coexistence, $L = 40$, $\mu^2 \approx 2.28$,
$\lambda = -1$, $\eta = 0.2$. \emph{Top:} representative configurations on
one shared symmetric scale; the barrier panel is a retained ladder state at
$q = q_{\rm cut}$, as well as a sample from both ordered and disordered phases. \emph{Bottom:} order parameters,
unsmoothed histograms. \method resolves all three $m$ branches and both $q$
phases; tempered SMC, at matched population and mutation
budget, occupies only the disordered one, and a single
preconditioned HMC chain from a prior draw remains in the disordered phase and never leaves it.}
\label{fig:phi6}
\end{figure}

\section{Conclusion}

In this work, we isolated the most attractive feature of the classical nested sampling algorithm for estimating marginal likelihoods: the quenched path of monotonically decreasing energy. Rather than trying to construct an MCMC mutation kernel that can efficiently follow this path in high dimensions, which is notoriously challenging, we demonstrate that the hard energy constraint can be replaced with a family of soft microcanonical potentials that preserves the quenched path while admitting efficient use of standard gradient-based mutation kernels. The resulting Quenched Ensemble Sampling method is a sequential Monte Carlo algorithm that performs at scale and is robust to first-order phase transitions, where tempering is not. In thermodynamic terms, the method is simulated quenching where tempering is simulated annealing, and the ability to follow this path of distributions at scale is unique.

Beyond Bayesian inference, estimating partition functions and free energies is a central problem in computational physics and chemistry. Flat-histogram methods, which explore the free-energy landscape by constructing an approximately uniform representation in energy, are widely used for this purpose~\citep{witman_flat-histogram_2018}. Nested sampling provides a related construction in which the energy levels adapt automatically and optimally to the enclosed prior volume, establishing the quenched path as a natural, and scalable, way to resolve the free-energy landscape.

\subsubsection*{Acknowledgements}
The author thanks Will Handley and Mike Hobson for extended discussions on nested sampling that shaped the foundations of this work. This work was supported by a Google Research Grant, and with computational support from the Google TPU Builders program. This work was supported by the UKRI Frontier Research Guarantee [EP/X035344/1].

\subsubsection*{Reproducibility statement}
The code developed for this work is available at \url{https://github.com/yallup/quenched_sampling}

\subsubsection*{AI Usage Statement}
The research code for this work was developed with the assistance of Claude (Fable 5), and was verified by the author against the closed-form analytic benchmarks reported in the paper. Generative AI tools (GPT 5.6 and Claude Fable) were used to edit and improve text originally written by the author, and to assist with running experiments and figure preparation. Generative AI was not used for research ideation or for interpreting the results. The author has reviewed all AI-assisted work and takes responsibility for the final content of this work, including text, claims or artifacts produced with the aid of generative AI.

\bibliography{quench,extra}
\bibliographystyle{iclr2026_conference}

\appendix
\crefalias{section}{appendix}
\crefalias{subsection}{appendix}
\section{Additional experimental details}
\label{app:experimental-details}

\subsection{Importance-sampling initialization}
\label{app:initialization}

Both QES and tempered SMC may be initialized by an initial sampling
step from the prior. For QES, we draw an anchor population of
$N_{\mathrm a}=10N$ independent samples $\widetilde x_i\sim\pi$ and choose the
initial energy threshold $E_0$ by bisection until the weights
$\omega_i^{(0)}=(E_0-U(\widetilde x_i))_+^\nu$ have ESS $N$. Their empirical
mean initializes the level volume,
\begin{equation}
  \widehat G_\nu(E_0)
  =\frac{1}{N_{\mathrm a}}\sum_{i=1}^{N_{\mathrm a}}\omega_i^{(0)},
  \label{eq:initial-volume}
\end{equation}
and their normalized values define an importance-sampling approximation to
$\rho_{E_0}$, from which the working population is resampled. This stabilizes
the initial importance step and prevents isolated outlying density evaluations
from determining the initialization.

Levels above $E_0$ may be estimated from the same prior draws and included in
the evidence quadrature. Tempered SMC can use the same construction by replacing
$\omega_i^{(0)}$ with the importance ratio from the prior to its chosen first
tempered target.

\subsection{Sampler configurations}
\label{app:config}

The configurations below are grouped by experiment because the population
size, target ESS and mutation budget are selected at the scale of each
problem. Within an experiment, QES and tempered SMC use the same population,
diagonally preconditioned MALA kernel, mutation budget and target ESS; only the
sequence of intermediate distributions differs. This ensures the experiments have a strong control that can
isolate
the effect of the quenched path. Unless stated otherwise, reported values are
averaged over three independent sampler seeds. QES uses softness $\nu=2$ throughout, and
triggers full resampling when the lineage-grouped ESS falls below $0.5N$
(\Cref{app:resample}). The diagonal preconditioner and scalar step size are
estimated from the preceding population and then held fixed during each
level's mutation, with the MALA acceptance rate targeted at $0.574$.

\paragraph{Analytic targets.}
All methods use $N=1000$ particles. QES and tempered SMC choose
successive levels at a default target ESS of $0.95$. The mutation budget
follows $\sqrt{D}$, the rate at which a random walk crosses a $D$-dimensional
shell, rounded up to a power of two so that ladder cost stays predictable and
floored at the value calibrated on the smallest targets,
\begin{equation*}
  n_{\mathrm{steps}}
  = \max\!\left(16,\; 2^{\left\lceil \log_2\!\sqrt{D}\,\right\rceil}\right),
\end{equation*}
MALA steps per level. Over the suite this takes four values: $16$ for
$D\le200$, $32$ at $D=500$ and $1000$, $64$ at $D=2000$, and $128$ at
$D=5000$ and $10^4$. The Gaussian run terminates at
$\dlogz=-3$. For the spike--slab and bimodal targets, the threshold is
deepened to $-2.5D-20$, where $D$ is the dimensionality of the target, this ensures that termination occurs beyond the transition rather than in
the broad phase. Whilst this seems like a problem-specific tuning, it is in reality a general requirement for any sampler to reach the low-energy phase of a first-order transition. When scanning for phase change behavior in e.g.~\Cref{sec:bnn}, the depth can be set arbitrarily low. On physical targets, the depth required can often be motivated from ground state energy estimates. As such $\dlogz =-3$ is more of a rule to match the efficiency of tempering on standard Bayesian posterior targets, rather than a sensitive tuning parameter, and in practice quenching can continue until all particles are at the ground state, although this would be inefficient for many problems.

The nested sampling baseline uses $1000$ particles,
removes $10\%$ of the live set per iteration, and applies $2D$ slice-sampling
steps per replacement. Its cost is reported in likelihood evaluations rather
than gradients; consequently, only scaling exponents, not absolute
intercepts, are compared across that boundary.

\paragraph{Bayesian neural networks.}
QES and tempered SMC use $N=1000$, target ESS $0.90$, and $96$ MALA steps per
level, with three sampler seeds for each fixed data split. NUTS uses $1000$
independent chains initialized from the prior, $250$ window-adaptation steps,
and $1000$ retained transitions. MCLMC follows the tuning procedure of
\citet{sommer_microcanonical_2024}, using $1000$ prior-initialized chains,
$2000$ tuning steps, and $10^4$ sampling steps. The deep ensemble optimizer initialisation is omitted as we are interested in purely sampling performance. MCLMC traces are thinned by a
factor of ten only when computing the rank-normalized ESS; all simulated steps
remain included in the gradient count. The Laplace--GGN baseline uses a MAP
estimate obtained with $2\times10^4$ full-batch Adam steps at learning rate
$10^{-3}$, followed by a generalized Gauss--Newton approximation. This
positive-semidefinite curvature is preferable to the raw Hessian for the
non-convex network posterior.

\paragraph{Lattice field theory.}
At the $40\times40$ coexistence point, QES and tempered SMC use $N=8000$,
target ESS $0.95$, and $64$ MALA steps per level. The termination threshold
$\dlogz=-850$ is chosen to extend the quenched ladder beyond the
approximately $741$-nat uniform-field barrier. Tempered SMC is additionally
run at target ESS $0.99$ and $0.999$ to test whether schedule refinement
recovers the ordered phase. The single-chain control uses window-adapted NUTS
with $500$ warmup steps and $2000$ retained states from a prior draw.

All samplers are implemented in JAX and BlackJAX and run in
\texttt{float32} on a single TPU v6e. Reported gradient totals
include adaptation and tuning; nested sampling instead counts every
likelihood evaluation.


\subsection{Problem definitions}
\label{app:targets}

\paragraph{Analytic targets.}
All three targets use the standard Gaussian prior
$\pi=\mathcal{N}(0,I_D)$ and a Gaussian-mixture likelihood, so their evidence
and posterior moments are available in closed form. The Gaussian and
spike--slab targets share the form
\begin{equation}
  e^{-U(x)}
  = \sum_{j=1}^{J} h_j
    \exp\!\left(-\frac{\lVert x\rVert^2}{2\sigma_j^2}\right).
  \label{eq:analytic-mixture}
\end{equation}
Writing $\tau_j^2=\sigma_j^2/(1+\sigma_j^2)$, their evidence and posterior
component weights are
\begin{equation}
  Z=\sum_{j=1}^{J}h_j\tau_j^D,
  \qquad
  \omega_j=\frac{h_j\tau_j^D}{Z}.
  \label{eq:analytic-mixture-evidence}
\end{equation}
Conditional on component $j$, the posterior is $\mathcal{N}(0,\tau_j^2I_D)$.
This gives an exact reference for both the evidence and posterior marginals.

\begin{description}
  \item[Gaussian] a single isotropic component ($J=1$, $\sigma=0.5$) over all $D$ coordinates. This smooth, unimodal
    target provides a scaling control without a phase transition.
  \item[Spike--slab] a broad component $\sigma_1=1$ and a narrow
    component $\sigma_2=0.1$. The narrow component's height is chosen so that
    it carries $90\%$ of the evidence at every $D$. The posterior phases occupy
    disjoint radial shells, producing a controlled first-order transition whose
    depth grows with dimension.
  \item[Bimodal] two equal-height anisotropic components centered at
    $\pm\mu$, with coordinate widths logarithmically spaced from $10^{-2}$ to
    $1$. The posterior modes are related by symmetry and therefore each carry
    exactly half of the evidence, making mode occupancy an exact diagnostic of
    population balance.
\end{description}

For the first two targets, $U$ depends on $x$ only through
$s=\lVert x\rVert^2$, with $s\sim\chi^2_D$ under the prior. Their
softened level volume therefore reduces to the one-dimensional integral
\begin{equation}
  G_\nu(E)
  = \int_0^{s_E} p_{\chi^2_D}(s)\,[E-U(s)]^\nu\,\mathrm{d}s,
  \label{eq:analytic-level-volume}
\end{equation}
where $s_E$ solves $U(s_E)=E$. Numerical quadrature of this expression
provides an independent reference for the level-volume estimate at every rung,
not only for the final $Z$. 

\paragraph{Bayesian neural networks.}
The UCI classification tasks use fully connected networks with two hidden layers of
width $16$. Unless an activation comparison is stated explicitly, the
activation is $\tanh$. Every parameter has an independent standard normal prior; each
layer's weight contribution is scaled by $1/\sqrt{d_{\mathrm{in}}}$ in the
forward pass and hidden-layer biases are scaled by $0.1$. Thus the effective
weights have variance $1/d_{\mathrm{in}}$ while the sampler retains isotropic
standard-normal coordinates. Inputs are standardized
using training-set statistics from a fixed 90--10 split. Classification uses
a categorical likelihood on the network logits $f_k$,
\begin{align}
  p(y \mid x, \theta)
  = \frac{\exp f_{y}(x, \theta)}{\sum_{k=1}^{K} \exp f_{k}(x, \theta)},
\end{align}
where $K$ is the number of classes. Reported test negative
log-likelihoods are the posterior-predictive mixture, evaluated over $S$
posterior draws $\{\theta_s\}$ and the held out test set $\{(x_i,y_i)\}$ of size $n_{\mathrm{test}}$,
\begin{align}
  \mathrm{NLL}
  = -\frac{1}{n_{\mathrm{test}}}
    \sum_{i=1}^{n_{\mathrm{test}}}
    \log \frac{1}{S} \sum_{s=1}^{S} p(y_i \mid x_i, \theta_s).
\end{align}

The datasets chosen correspond to the Red Wine Quality~\citep{paulo_cortez_wine_2009}, Heart Disease~\citep{andras_janosi_heart_1989}, Australian Credit Approval~\citep{quinlan_statlog_1987}, Sonar~\citep{terry_sejnowski_connectionist_1988} and Glass identification~\citep{b_german_glass_1987} tasks from the UCI repository. These are small classification tasks, with a mixture of continuous and discrete inputs from input dimension 9-60, and a mixture of binary and multi-class outputs. The dataset sizes range from 208-1599, generally small enough to allow for rapid experimentation with a full batch likelihood. Scaling to datasets with larger number of inputs typically requires employing Stochastic Gradient MCMC techniques~\citep{nemeth_stochastic_2021} such as SGLD~\citep{welling_bayesian_2011}. How well these approaches perform relative to full-batch methods is an open question~\citep{wenzel_how_2020}, and employing a quenched path in a mini-batch likelihood settings is conceptually challenging. Comparison of marginal likelihood between full-batch quenched path estimation and exhaustive leave $n$-out cross-validation would be an interesting future direction~\citep{fong_marginal_2020}.

\paragraph{Lattice field theory.}
The lattice target is the periodic $L=40$ scalar field defined in
\Cref{eq:phi6}, with an exactly sampled Gaussian free-field prior at
$m_0^2=1$. We set $\lambda=-1$, $\eta=0.2$, and use the finite-volume
coexistence point $\mu^2=2.28347$, located by bracketing pilot QES ladders at
$L=40$ on equal phase weight; the equal-weight points measured this way at
$L=8$--$40$ drift with the expected $2\ln 2/(V\,\Delta q)$ multiplicity
correction, extrapolating to $\mu^2_\infty\approx2.283$. Both particle
methods include the exact
$Z_2$ transformation $\phi\mapsto-\phi$ as an invariant move, removing sign
switching as a confound. The phase coordinate
$q=L^{-2}\sum_x\phi_x^2$ is invariant under this move, so crossing between the
ordered and disordered phases must still be achieved by the sampling path.

\section{Mobility in the soft microcanonical family}
\label{app:nu}

\begin{figure}[H]
\begin{center}
\includegraphics[width=\textwidth]{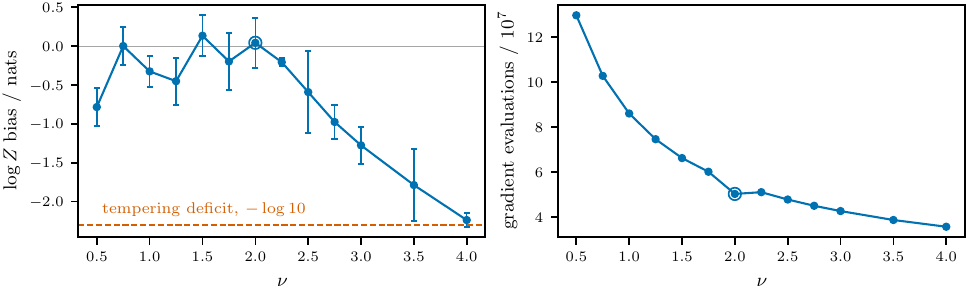}
\end{center}
\caption{Softness sweep on the spike--slab target at $D=200$, with three
seeds per point and the default $\nu=2$ circled. \emph{Left:} evidence bias;
the dashed line marks the $-\log 10$ deficit of tempered SMC, corresponding to
loss of the spike carrying $90\%$ of the evidence. \emph{Right:} total
gradient evaluations. Small $\nu$ approaches the hard-constraint limit and
lengthens the ladder, whereas large $\nu$ lowers the cost but becomes
tempering-like and misses the spike.}
\label{fig:nu}
\end{figure}

The exponent $\nu$ controls the strength at which the barrier repulsion appears. As illustrated in \Cref{fig:nu}, the effective gradient of the boundary term becomes a pronounced steep force close to the boundary, a valid concern is then how \emph{mobile} a particle can be in practice under this force. 
Writing the slack from the boundary as $s=E-U(x)$, the contribution of the
soft constraint to the score is
\begin{equation}
  \nabla\log (E-U(x))_+^\nu
  = -\frac{\nu}{s}\nabla U(x).
  \label{eq:wall-score}
\end{equation}
As $\nu\to0$, the target approaches the hard indicator used by nested
sampling. At the opposite extreme, setting $E=\nu/\beta$ and letting
$\nu\to\infty$ gives
$(E-U)_+^\nu = E^\nu\,(1-\beta U/\nu)_+^\nu \to E^\nu e^{-\beta U}$, so a
joint scaling of level and softness recovers the canonical ensemble at
inverse temperature $\beta$: the family interpolates exactly between the
nested-sampling constraint and tempering. We aim to motivate the specific choice $\nu=2$ as a compromise between the two extremes.

Firstly, we note that the choice $\nu=2$ has a useful boundary interpretation. Let
$\beta_E=\partial_E\log g(E)$ denote the local slope of the density of states.
When $\log g$ is approximately linear over the typical slack, $g(E-s)\simeq g(E)e^{-\beta_Es}$ and hence
\begin{equation}
  s\mid E \;\simeq\; \mathrm{Gamma}(\nu+1,\text{rate }\beta_E),
  \qquad
  \mathbb{E}\!\left[\frac{\nu}{s}\right]\simeq\beta_E,
  \qquad
  \operatorname{Var}\!\left[\frac{\nu}{s}\right]
  \simeq\frac{\beta_E^2}{\nu-1}.
  \label{eq:wall-moments}
\end{equation}
The variance is finite for $\nu>1$, making $\nu=2$ the smallest integer for
which both the mean and variance of the boundary contribution are finite. At
the same time, the target density vanishes quadratically at $U=E$, replacing
the hard wall by a continuously differentiable turning surface. Whilst there may be stronger arguments for other values of $\nu$, we primarily motivate the choice empirically, using the spike--slab target to illustrate the trade-off between bias and cost. 

\paragraph{Selecting the softness.}
We sweep $\nu$ on the $D=200$ spike--slab target, where a genuine first-order
transition makes excessive departure from the nested sampling limit to tempering directly observable. All
other settings are held fixed, and each point is repeated over three seeds.

The accurate region is broad: every point in
$0.75\lesssim\nu\lesssim2.25$ is consistent with the analytic evidence within
the seed variation. At the hard end, the gradient cost increases by a factor
of $3.6$ between $\nu=4$ and $\nu=0.5$, and the estimate at $\nu=0.5$ is
$-0.78$\pmerr{0.24} nats low. Above $\nu\simeq2.5$, the bias moves steadily
toward the tempered-SMC deficit, reaching $-2.24$\pmerr{0.09} nats at
$\nu=4$. The default $\nu=2$ is consequently near the least expensive edge of
the accurate region while remaining separated from the tempering-like regime.
Together with the finite score variance in~\Cref{eq:wall-moments}, this gives a
single choice that requires no target-specific adjustment in our experiments.

\paragraph{Mobility near the boundary.}
We measure actual particle movement on the isotropic Gaussian, defining a
proposal as near the boundary when its slack is below one fifth of the mean
slack at that level. With $\nu=2$, only $3$--$7\%$ of proposals fall in this
region. Their move rate is lower than in the bulk, but remains substantial:
at $D=1000$ it is $0.289$, compared with $0.598$ in the full population. The
two-dimensional case is the least mobile, and the near-boundary to overall ratio
increases from $0.31$ at $D=2$ to $0.65$ at $D=3$ before remaining of order
one half at larger dimensions. Thus the soft boundary slows a small fraction
of walkers rather than creating a frozen layer, and the effect does not worsen
toward zero mobility as dimension increases.

The local Gamma approximation in~\Cref{eq:wall-moments} predicts that $2.31\%$
of an equilibrated $\nu=2$ population lies below this slack threshold. The
larger measured fraction reflects finite relaxation after each level change:
lowering $E$ reduces every particle's slack simultaneously, after which the
mutation kernel moves the population back into the interior. 

\begin{table}[H]
\centering
\begin{tabular}{@{}lrrrr@{}}
\toprule
$D$ & near the wall & moves there & moves overall & ratio \\
\midrule
$2$ & $3.6\%$ & $0.126$ & $0.403$ & $0.31$ \\
$3$ & $3.2\%$ & $0.282$ & $0.434$ & $0.65$ \\
$5$ & $3.5\%$ & $0.327$ & $0.464$ & $0.70$ \\
$10$ & $4.3\%$ & $0.366$ & $0.497$ & $0.74$ \\
$50$ & $6.1\%$ & $0.286$ & $0.554$ & $0.52$ \\
$1000$ & $7.1\%$ & $0.289$ & $0.598$ & $0.48$ \\
\bottomrule
\end{tabular}

\caption{Mobility near the level-set boundary on the isotropic Gaussian, with
$N=200$, $n_{\mathrm{steps}}=10$, and $\nu=2$. ``Near the wall'' denotes
slack below one fifth of the level mean. The two move columns report the
fraction of proposals that change the particle position, measured directly
from the trajectories; the final column is their ratio.}
\label{tab:wall}
\end{table}

\section{Accounting and posterior reconstruction}
\label{sec:unbiased}
\label{app:accounting}

The evidence estimate, posterior reconstruction, resampling schedule and
mutation budget are different parts of one sequential construction. This
section separates their roles. The central distinction is between quantities
that are \emph{measured} from the weighted population and operations that only
change how that population is represented.

\subsection{Assigned and measured level volumes}
\label{app:measured-volumes}

Both nested sampling and SMC estimate the normalising constant from a series of incremental ratios along a
path. Classic nested sampling assigns the compression
$t_k=G_k/G_{k-1}$ from an order-statistic law: if the $m$ live points are
independent draws from the constrained prior, their enclosed masses are
independent $\mathrm{Uniform}[0,1]$ variables, and discarding the worst point
gives $t_k\sim\mathrm{Beta}(m,1)$. Energies are measured, but volumes are
deduced from this law~\citep{skilling_nested_2006,chopin_properties_2010}.

QES instead measures every soft-volume increment from the population present
at the level, following a more standard SMC
approach. For particles with normalised
weights $w_j$ targeting $\rho_{E_k}$, the pointwise ratio $r$
of~\Cref{eq:ratio}, evaluated for the pair $(E_k, E_{k+1})$, gives
\begin{equation}
  \widehat r_k
  = \sum_{j=1}^N w_j\, r(x_j),
  \qquad
  \widehat G_{\nu,k+1}
  = \widehat G_{\nu,k}\widehat r_k,
  \label{eq:appendix-measured-ratio}
\end{equation}
where the ratio is evaluated before mutation. The ratio is of unnormalised
densities: the normalised ratio $\rho_{E_{k+1}}/\rho_{E_k}$ has expectation
one under $\rho_{E_k}$, since the normalisers are the quantity being
estimated. This is the standard SMC
normalising-constant construction applied to the soft level
family~\citep{salomone_unbiased_2025}.

The distinction changes the assumptions, and resulting guarantees of the sampling process. For a fixed, externally specified
schedule, exact incremental weights and mutation kernels invariant for their
current targets give an unbiased unnormalised evidence estimator for any
particle count and any amount of mutation
\citep{del_moral_feynman-kac_2004}. Poor mutation results in increased variance, without adding a mixing bias to that estimator. Normalised posterior expectations and
$\log\widehat Z$ still have their usual finite-particle bias. The classic nested
sampling assignment of volumes carries a stricter requirement, namely that the live particles must also
be mutually independent. A replacement chain that remains correlated with
its surviving start violates the order-statistic premise, so the assigned
compressions need not describe the realised population. In practice the bias is often small, but practical implementations of nested sampling often use relatively long mutation chains to keep this in check. As a result, when it comes to efficient scaling with a matched comparison to tempered SMC, as in this work, we find the more standard SMC accounting to be more appropriate.

As a result, QES retains the vertical path but uses the measured convention. Killing a
particle at a new level gives it incremental weight zero, so its mass remains
in the denominator of~\Cref{eq:appendix-measured-ratio} and is automatically
banked in $\widehat G_{\nu,k}-\widehat G_{\nu,k+1}$. Replacing a particle of
weight $w$ by two coincident copies of weight $w/2$ also leaves the empirical
measure exactly unchanged,
\begin{equation}
  \frac{w}{2}\delta_x+\frac{w}{2}\delta_x=w\delta_x.
\end{equation}
Mutation subsequently moves the copies apart. We find that while the population diversity is sufficiently high, the simple particle splitting is efficient, and the systematic resampling step is only triggered when the ESS of the carried weights falls below a threshold. This is discussed in more detail in \Cref{app:resample}.

\subsection{Adaptive levels and finite-particle bias}
\label{app:adaptive-levels}

The next energy level is selected from the current population by solving for
the value of $E_{k+1}<E_k$ at which the prospective incremental weights reach
the target ESS. This is an adaptive schedule: it avoids specifying the unknown
shape of $E(G)$ in advance and automatically shortens the step when the
population would otherwise lose too much support. It is also the one departure
from the fixed-schedule unbiasedness result above. Adaptive SMC estimators
remain consistent and satisfy a central limit theorem
\citep{beskos_convergence_2016}. The method therefore preserves the practical
adaptivity of nested sampling without claiming exact finite-$N$ unbiasedness.

This level-selection ESS is distinct from the second ESS used to decide when
to redraw the population. The first sets the distance between adjacent
targets; the second controls accumulated weight degeneracy across many such
steps. Conflating them would force a full resampling at every rung of a path
whose length grows rapidly with dimension.

\subsection{Pooling the path into a posterior sample}
\label{app:pooling}
\label{app:typical}

The same accounting reconstructs the posterior from the entire path, rather
than only from its final cloud. From~\Cref{eq:soft-evidence}, level $k$ carries
quadrature weight
\begin{equation}
  b_k \;\propto\;
  \widehat G_\nu(E_k)e^{-E_k}\,\Delta E_k.
\end{equation}
Here $\Delta E_k$ is the positive finite-difference width computed from the
stored descending ladder. If $w_{kj}$ is particle $j$'s normalised carried
weight at that level, then the pooled posterior weight is
$q_{kj}\propto b_k w_{kj}$. We report the Kish quantity
\begin{equation}
  N_{\mathrm{Kish}}
  = \frac{\bigl(\sum_{k,j}q_{kj}\bigr)^2}
         {\sum_{k,j}q_{kj}^2}
  \label{eq:pooled-kish}
\end{equation}
over every particle of every rung. For tempered SMC, let
$\beta_1,\ldots,\beta_K$ denote the recorded post-mutation stages. Because
each stage has the same population size, a particle at stage $\beta_k$ is
reweighted against the equally weighted deterministic mixture of stages,
\begin{equation}
  q_{kj}\;\propto\;
  \frac{e^{-U(x_{kj})}}
       {\sum_{\ell=1}^K \widehat Z_{\beta_\ell}^{-1}
        e^{-\beta_\ell U(x_{kj})}},
\end{equation}
where $\widehat Z_{\beta_\ell}$ is the normalising-constant estimate
accumulated by that SMC run. The prior density and the common mixture factor
cancel from this ratio. This means both methods are scored in efficiency using all particles in all
recorded clouds. We form this complete weight vector and evaluate
\Cref{eq:pooled-kish} once per run; ESS per gradient is computed for that run
before results are averaged over repeats. 

For the particle methods, pooled Kish ESS measures the concentration of this
full posterior-weight vector across particles and levels. It does not account
for shared ancestry, mutation correlation or
dependence between neighbouring levels, and is therefore a nominal rather than
correlation-adjusted sample size. For NUTS and MCLMC, bulk ESS accounts for
within-chain correlation. Both diagnostics indicate how efficiently
computation is converted into posterior samples and are qualitatively
comparable, but their values should not be interpreted as exact ratios of
independent draws.

The large QES pool follows directly from the geometry of the energy path. For
the isotropic Gaussian, let $\pi=\mathcal N(0,I_D)$ and
$U(x)=\lVert x\rVert^2/(2\sigma^2)$. The posterior is
$\mathcal N(0,\sigma_\star^2I_D)$ with
$\sigma_\star^2=\sigma^2/(1+\sigma^2)$. Under the exact augmentation,
$E=U+s$, where $s\sim\mathrm{Gamma}(\nu+1,1)$ is independent of $x$, so
\begin{equation}
  \operatorname{Var}(E)
  = \frac{D}{2(1+\sigma^2)^2}+\nu+1.
  \label{eq:varE}
\end{equation}
The posterior-weighted range of levels therefore has width
$\Theta(\sqrt D)$.

At a typical level the local form
$g(E-s)\simeq g(E)e^{-\beta_Es}$ implies that the moments of the incremental
weight depend on the level drop through $\beta_E\Delta E$. A fixed
incremental ESS consequently fixes $\beta_E\Delta E$; for this target
$\beta_E\to1$, so $\Delta E=\Theta(1)$. The full QES path spans an
$\Theta(D)$ energy range and therefore contains $\Theta(D)$ rungs, of which
$\Theta(\sqrt D)$ carry posterior weight.

Tempering has the complementary scaling. Its incremental ESS fixes
$\Delta\beta=\Theta(D^{-1/2})$, giving $\Theta(\sqrt D)$ stages across
$\beta\in[0,1]$. A stage can be reweighted to the posterior without weight
collapse only when $1-\beta=O(D^{-1/2})$, so only $\Theta(1)$ stages carry
posterior weight. Hence, before accounting for within-level weight variation,
the two pooled Kish counts scale as
\begin{equation}
  N_{\mathrm{Kish}}^{\mathrm{QES}}=\Theta(N\sqrt D),
  \qquad
  N_{\mathrm{Kish}}^{\mathrm{SMC}}=\Theta(N).
\end{equation}
This is a path-level sample-size advantage, not a free computational
advantage: QES also uses $\sqrt D$ more stages, cancelling the additional
$\sqrt D$ in its pooled Kish. At fixed mutation cost per stage, both methods
therefore scale as $\Theta(D^{-1/2})$ Kish per gradient. Under the benchmark
budget $n_{\mathrm{steps}}=\Theta(\sqrt D)$, both acquire the same additional
$D^{-1/2}$ cost and scale as $\Theta(D^{-1})$.

\Cref{tab:typical} verifies these counts. Across $D=10$ to $10^4$, the
number of QES rungs carrying posterior weight grows from $20.0$ to $674.9$,
while the tempered count remains between $9.5$ and $9.9$. The full QES ladder
grows from $43$ to $14{,}493$ rungs, whereas the tempered path grows from $16$
to $502$ stages.

\begin{table}[H]
\centering
\begin{tabular}{@{}lrrrr@{}}
\toprule
 & \multicolumn{2}{c}{\method} & \multicolumn{2}{c}{tempered SMC} \\
\cmidrule(lr){2-3}\cmidrule(l){4-5}
$D$ & rungs & carrying weight & stages & carrying weight \\
\midrule
$10$ & $43$ & $20.0$ & $16$ & $9.9$ \\
$50$ & $139$ & $48.8$ & $35$ & $9.5$ \\
$100$ & $238$ & $68.9$ & $50$ & $9.7$ \\
$200$ & $416$ & $93.7$ & $71$ & $9.8$ \\
$500$ & $892$ & $144.4$ & $112$ & $9.5$ \\
$1000$ & $1684$ & $217.3$ & $158$ & $9.7$ \\
$2000$ & $3144$ & $303.6$ & $224$ & $9.5$ \\
$5000$ & $7446$ & $488.6$ & $354$ & $9.6$ \\
$10000$ & $14493$ & $674.9$ & $502$ & $9.9$ \\
\bottomrule
\end{tabular}

\caption{Path length and posterior-weight-carrying stages on the isotropic
Gaussian, with $N=1000$, three seeds and incremental ESS target $0.95$.
``Carrying weight'' is the Kish quantity formed from the total weight of each
rung or stage. For QES this count grows as $\sqrt D$ while the full ladder
grows as $D$; for tempered SMC the full path grows as $\sqrt D$ while the
weight-carrying count remains bounded.}
\label{tab:typical}
\end{table}

\subsection{Resampling cadence}
\label{app:resample}

The long QES ladder makes resampling cadence more of an issue, particularly for multimodal targets. Redrawing at every
rung gives fresh uniform weights, but it also repeatedly resamples neutral
mode labels. On a symmetric multimodal target with no local transitions,
occupancy then performs a random walk with absorbing states at zero and one.
At $D=100$, four of ten runs fixate during the approximately $1{,}800$
per-level redraws, even though the measured death rates are identical in the
two modes ($0.0040$ in each).

At the other extreme, branching can replace zero-weight particles while all
survivor weights are carried indefinitely. This preserves the represented
measure, but eventually concentrates it on few ancestral lineages, increasing
variance and the finite-$N$ bias of nonlinear summaries such as
$\log\widehat Z$. We therefore branch at every level and fully resample only
when the carried-weight ESS falls below $fN$~\citep{moral_adaptive_2012}. The trigger is computed after
grouping weights by lineage, splitting a donor into two children halves its
contribution to the ordinary sum of squared weights and would spuriously
increase the ungrouped ESS, although it has created no new independent
lineage. With the default $f=0.5$, full resampling occurs about every $14$
levels at an incremental ESS target of $0.95$, roughly fourteen times less
often than per-level resampling.

\begin{figure}[H]
\centering
\includegraphics[width=\textwidth]{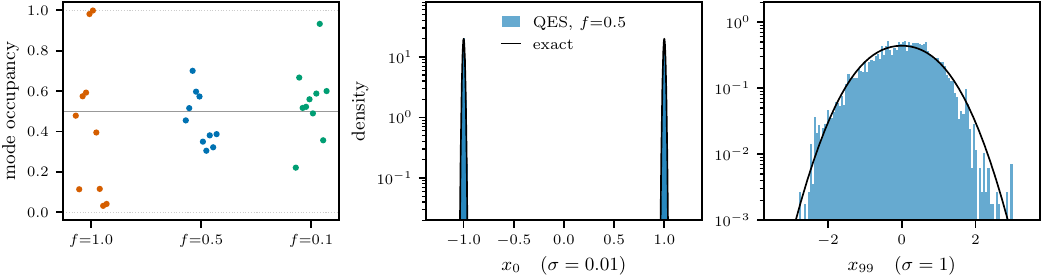}
\caption{\emph{Left:} mode occupancy for every run at each resampling
threshold, ten seeds per setting on the bimodal target at $D=100$; a fixated
run lies on a dotted boundary. Per-level resampling ($f=1$) drives neutral
occupancy toward fixation, whereas the default $f=0.5$ preserves both modes in
every run. \emph{Middle, right:} posterior marginals under the default against
the exact density at the two ends of the anisotropy, $\sigma=0.01$ and
$\sigma=1$.}
\label{fig:cadence}
\end{figure}

\Cref{fig:cadence} illustrates this trade-off on the bimodal target
at $D=100$. Resampling every level causes the occupancy to collapse into one
mode, while the default cadence preserves both modes in every run and
reconstructs the marginals at both ends of the anisotropy. Resampling less
often than the default lets the occupancy drift back toward imbalance, with
one of ten runs approaching fixation. Accounting for this drift is a
challenge for any interacting particle method, and motivates global moves
that cross between modes; constructing such moves is itself hard in high
dimension without an \emph{a priori} known symmetry.

\subsection{Mutation budget}
\label{app:budget}

\begin{figure}[H]
\centering
\includegraphics[width=0.72\textwidth]{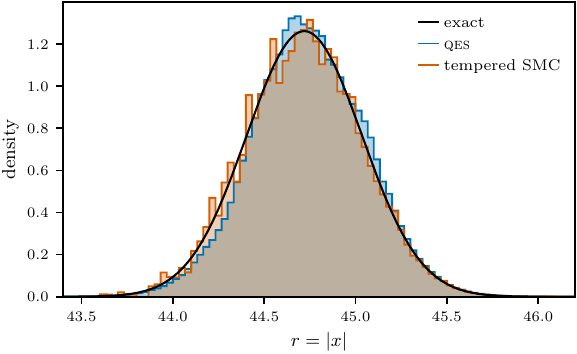}
\caption{Reconstructed radial posterior at $D=10^4$ against the exact shell
$r=\tau\chi_D$, for a single run at the default working point. Both methods
pool every weighted particle from every level of the path: \method with the
level-volume quadrature weights of \Cref{app:pooling}, tempered SMC with
deterministic-mixture balance weights over its temperatures. The pooled
Kish ESS is $4.8\times10^{5}$ for \method{} and $6.9\times10^{3}$ for
tempered SMC.}
\label{fig:rfill}
\end{figure}

The residual Gaussian bias at $D=10^4$ (\Cref{tab:analytic}) is probed by
sweeping the number of MALA steps per level around the working point of
$n_{\mathrm{steps}}=128$, with every other setting of the benchmark held
fixed. Below the default the bias grows steadily as the population lags its
moving target, while doubling the budget changes nothing within the seed
variation (\Cref{tab:nsteps}). The default therefore sits at the knee of the
curve, and we attribute the residual that remains there to the accumulated
finite-$N$ effects of \Cref{app:accounting} rather than to incomplete
mutation. Doubling the particle count at the default budget instead shrinks
the bias and contracts the seed scatter by the predicted $\sqrt{2}$,
consistent with this attribution. \Cref{fig:rfill} shows the reconstructed
radial posterior at the default working point.

\begin{table}[H]
\caption{Budget sweeps on the Gaussian at $D=10^4$: evidence bias against
the exact $\log Z$ and gradient evaluations per run, ten seeds per setting.
The default is $N=1000$, $n_{\mathrm{steps}}=128$; the final row doubles the
particle count at the default mutation budget.}
\label{tab:nsteps}
\begin{center}
\begin{tabular}{@{}rrrr@{}}
\toprule
$N$ & $n_{\mathrm{steps}}$ & gradients & $\Delta \log Z$ \\
\midrule
$1000$ & $32$ & $0.47\times10^9$ & $-1.91$\pmerr{0.68} \\
$1000$ & $64$ & $0.94\times10^9$ & $-1.62$\pmerr{0.93} \\
$1000$ & $128$ & $1.86\times10^9$ & $-0.89$\pmerr{0.71} \\
$1000$ & $256$ & $3.69\times10^9$ & $-1.10$\pmerr{0.77} \\
\midrule
$2000$ & $128$ & $3.74\times10^9$ & $-0.62$\pmerr{0.45} \\
\bottomrule
\end{tabular}

\end{center}
\end{table}

\end{document}